\documentclass[sigconf,nonacm]{acmart}
\usepackage{placeins}
\usepackage{array}        
\usepackage{tabularx}     
\usepackage{booktabs}     
\usepackage{caption}      
\newcolumntype{Y}{>{\centering\arraybackslash}X} 
\usepackage{xcolor}
\usepackage{graphicx}     
\usepackage{subcaption}   
\usepackage{float}        
\usepackage{enumitem}
\usepackage{pifont}      
\definecolor{editpurple}{RGB}{153, 50, 204} 

\PassOptionsToPackage{draft}{graphicx}
\AtBeginDocument{%
  }

\setcopyright{none}
\copyrightyear{2026}
\acmYear{2026}
\acmDOI{}  
\acmConference[KDD '26]{Proceedings of the 2026 ACM SIGKDD Conference on Knowledge Discovery and Data Mining}{August 9--13, 2026}{Jeju, Republic of Korea}
\acmBooktitle{Proceedings of the 2026 ACM SIGKDD Conference on Knowledge Discovery and Data Mining (KDD '26), August 9--13, 2026, Jeju, Republic of Korea}
\acmISBN{}

\begin{document}
\title{M-GLC: Motif‐Driven Global‐Local Context Graphs for Few-shot Molecular Property Prediction}

\author{Xiangyang Xu}
\affiliation{%
  \institution{Iowa State University}
  \city{Ames}
  \country{USA}}
\email{xyxu@iastate.edu}

\author{Hongyang Gao}
\affiliation{%
  \institution{Iowa State University}
  \city{Ames}
  \country{USA}}
\email{hygao@iastate.edu}

\renewcommand{\shortauthors}{Xu et al.}

\begin{abstract}


Molecular property prediction (MPP) is a cornerstone of drug discovery and materials science, yet conventional deep learning approaches depend on large labeled datasets that are often unavailable. 
Few‐shot Molecular property prediction (FSMPP) addresses this scarcity by incorporating relational inductive bias through a context graph that links molecule nodes to property nodes, but such molecule-property graphs offer limited structural guidance.
We propose a comprehensive solution: Motif-Driven Global-Local Context Graph for few-shot molecular property prediction, which enriches contextual information at both the global and local levels. 
At the global level, chemically meaningful motif nodes representing shared substructures, such as rings or functional groups, are introduced to form a global tri-partite heterogeneous graph, yielding motif-molecule-property connections that capture long‐range compositional patterns and enable knowledge transfer among molecules with common motifs. 
At the local level, we build a subgraph for each node in the molecule-property pair and encode them separately to concentrate the model’s attention on the most informative neighboring molecules and motifs.
Experiments on five standard FSMPP benchmarks demonstrate that our framework consistently outperforms state‐of‐the‐art methods.
These results underscore the effectiveness of integrating global motif knowledge with fine‐grained local context to advance robust few‐shot molecular property prediction.

\end{abstract}



\maketitle

\section{Introduction}
\begin{figure*}[t!]
    \centering
    \includegraphics[width=1\linewidth,clip,trim=0 185 0 40]{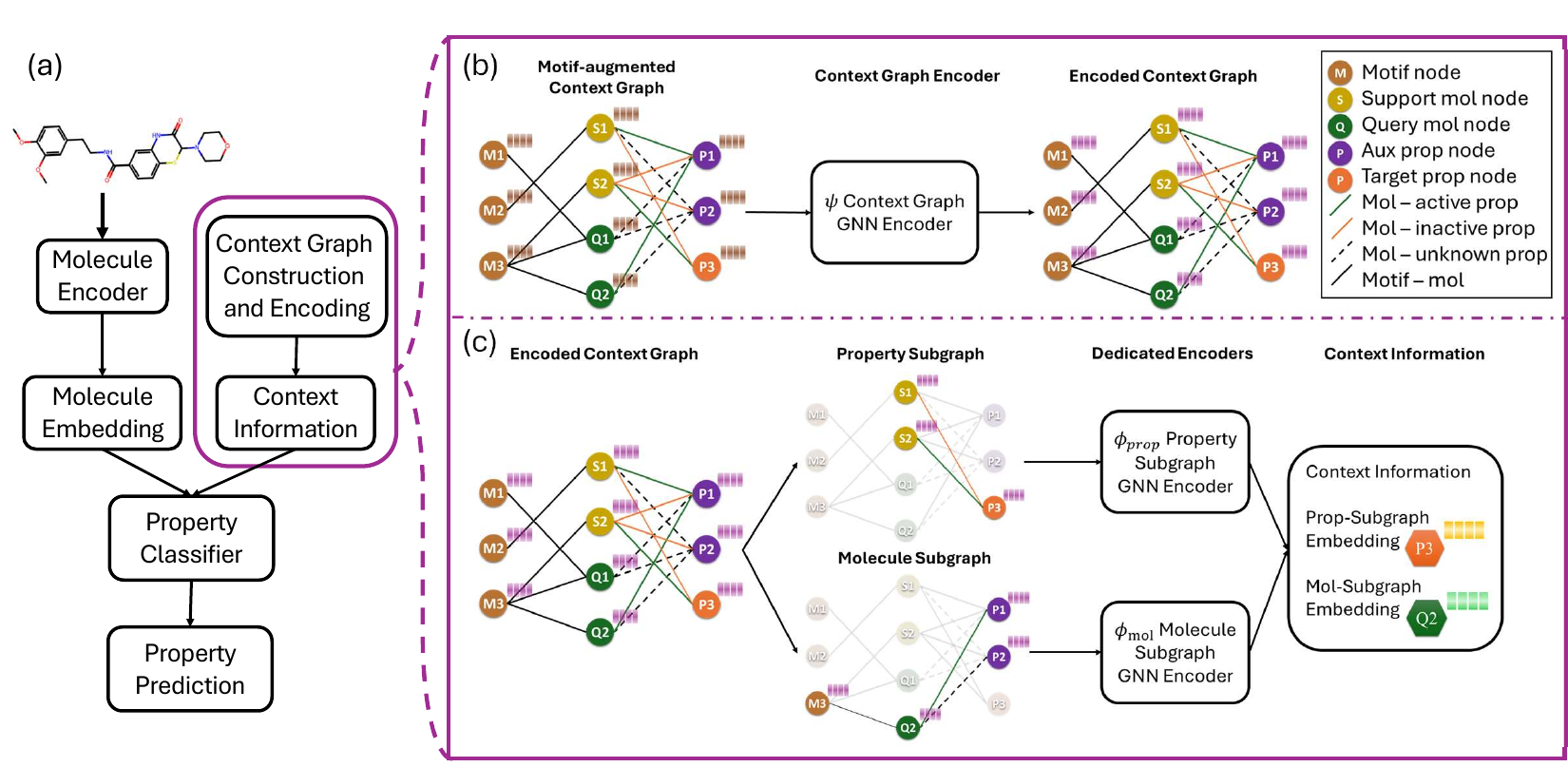}
    \Description{}
    \caption{
    (a) Framework of \textbf{M-GLC}: \textbf{M}otif-driven \textbf{G}lobal–\textbf{L}ocal \textbf{C}ontext for few-shot molecular property prediction. 
    Our work focuses on extracting informative context signals.
    (b) Firstly, we construct a tri-partite heterogeneous context graph with motifs, molecules, and properties to inject domain-specific structural priors. A global GNN encoder $\psi$ processes the full context graph to produce node embeddings that summarize global relational structure. 
    (c) Then, we select task-relevant \emph{local-focus} subgraphs centered on the target molecule and its target property. Finally, the dedicated subgraph encoders $\phi_{\text{mol}}$ and $\phi_{\text{prop}}$ map molecule-centric and property-centric subgraphs to subgraph-level embeddings, yielding a cleaner and more stable context representation for fast adaptation.
    }
    \label{frame}
\end{figure*}

Molecular Property Prediction (MPP) is a key task in drug development and materials science, aiming to predict the properties of molecules, such as toxicity and bioactivity \citep{fang2022geometry, sliwoski2014computational, riniker2013similarity, hansen2015machine}.
In the past few years, deep learning methods have been widely used for molecular property prediction \cite{dlmpp1, dlmpp3, dlmpp5, dlmpp6}. These methods usually use encoders such as graph neural networks (GNNs) to obtain molecular representations and then use classifiers to predict molecular properties. However, their performance usually relies on large-scale, fully labeled datasets for effective training.
Unfortunately, unlike the abundant and easily accessible data in Natural Language Processing (NLP) and Computer Vision (CV), labeled molecular data are limited in size and expensive to obtain. Molecular property labels usually rely on expensive wet-lab experiments \citep{nguyen2020meta,altae2017low,chen2022meta}, which makes large-scale datasets impractical in real-world scenarios.

Recent Few-Shot Molecule Property Prediction (FSMPP) studies \citep{schimunek2023context, wang2024pin, wang2021property, zhuang2023graph} have shown that \textit{context information}, the molecular embeddings learned from a molecule–property graph, plays an important role in few-shot molecular property prediction. These embeddings help capture the underlying relationships between molecules and tasks, thereby enabling better performance under few-shot learning settings.
Despite recent progress, current FSMPP methods still face several limitations. 

\textbf{(1)} Current molecular–property context graphs remain limited by label sparsity. Many mol–prop associations are missing even in standard benchmarks, which weakens the informativeness of the topology. In the few-shot learning setting, richer context is crucial to stabilize fast adaptation and reduce variance under scarce supervision. 
\textbf{(2)} Motif-level information is often either overlooked \citep{schimunek2023context, wang2024pin, wang2021property, zhuang2023graph} or incorporated through overly simplified strategies \citep{meng2023meta}, even though previous studies in MPP \citep{zhang2021motif, yu2022molecular} have demonstrated that effectively leveraging motif information can significantly enhance molecular representations and prediction performance.   
\textbf{(3)} The output embeddings of the heterogeneous context graph are typically node-level features extracted by a simple GNN encoder. However, node-level embeddings often struggle to capture critical structural characteristics and patterns \citep{alsentzer2020subgraph, sun2021sugar, ullmann1976algorithm}, and a simple GNN often fails to capture context information without representation entanglement \citep{zhou2024slotgat}.

To address these limitations, we propose \textbf{M-GLC}: \textbf{M}otif-Driven \textbf{G}lobal–\textbf{L}ocal \textbf{C}ontext Graphs, a novel comprehensive solution for few-shot molecular property prediction that integrates a tri‑partite context graph with structure‑aware aggregation and a local‑focus subgraph encoder.

\textbf{(1)} We construct a heterogeneous context graph with three node types: \textit{molecule}, \textit{motif}, and \textit{property}. This tri-partite structure retains the advantage of traditional context graphs, enabling effective label supervision through molecule–property associations. At the same time, the graph is enriched by introducing motif nodes, which provide connections based on shared chemical substructures, such as rings and functional groups. 
These motif-based connections supply additional structural information and serve as bridges across tasks, which is especially helpful when labeled data is limited. This enriched context allows the model to leverage both label signals and structural priors, improving generalization in few-shot settings.
\textbf{(2)} Instead of reading out node-level embeddings from a simple GNN only, we propose a Local-Focus Subgraph module that computes subgraph-level embeddings from the context graph. This design enhances the representation by focusing on relevant local patterns rather than a single node. In addition, dedicated subgraph encoders more efficiently capture context information from specific kinds of nodes.

The key contribution of our work can be summarized as follows:
\begin{itemize}[leftmargin=*,noitemsep,topsep=0pt]
\item We propose the tri-partite context graph that uses molecules, properties, and motifs as nodes, which is specifically designed to integrate motif-level structural information into context-based few-shot learning, enabling the model to capture both task-specific label signals and transferable structural priors.
\item We introduce the structure-aware edge-weighted aggregation to address the signal imbalance between motifs and molecules.
\item We introduce the subgraph-level context embeddings in place of node-level context embeddings, enabling the model to better capture complex structural patterns.
\item Experiments on five widely-used few-shot molecule property prediction benchmarks show that our method significantly outperforms previous state-of-the-art approaches.
\end{itemize}

\section{Related Work}
\textbf{Few-shot Molecular Property Learning}
FSMPP aims to train accurate molecular classifiers using only a small number of labeled examples per task. Early work \citep{altae2017low} applied metric- and meta-learning techniques without incorporating molecular structure. Later methods introduced graph neural networks (GNNs) \citep{nguyen2020meta, chen2022meta} to enable structure-aware learning in the few-shot setting.
To improve generalization, some approaches \citep{hu2019strategies, yang2021graphformers} adopted pretrained molecular encoders to provide transferable features. Combining pretraining with meta-learning has shown clear advantages over training from scratch \citep{guo2021few}.
More recent efforts \citep{wang2021property, zhuang2023graph, wang2024pin} go beyond individual molecules and model cross-task relationships using context graphs built from molecular similarities or auxiliary property signals. However, most of these methods rely on surface-level patterns and overlook common structural fragments, which often reflect functional roles.

\textbf{Structure-Aware Molecular Representations}  
Structure-aware molecular representations aim to enhance molecular modeling by explicitly integrating structural patterns beyond individual atoms. Early models \citep{gilmer2017neural, guo2020graseq, hu2019strategies, lu2019molecular} treated molecules as planar graphs, focusing on local atomic interactions while ignoring recurring substructures.
Later methods \citep{frag1-wu2024t, frag1-zhang2021fragat} incorporated predefined fragments or subgraphs as additional input, helping the model identify common functional groups. Hierarchical architectures \citep{frag2-zhu2022hignn, frag2-zang2023hierarchical} were proposed to represent molecules at multiple levels of resolution, allowing information to be aggregated from both fine-grained and coarse-grained perspectives. Some approaches \citep{wu2023molformer, yu2022molecular} further embed substructures as separate nodes in extended graphs, enabling direct interaction between motifs and atoms. These enhancements improve generalization and robustness across property prediction tasks. However, existing methods require large labeled datasets to train structure-aware encoders, limiting their usage in low-resource settings.

\section{Preliminary}
\subsection{Few-Shot Molecular Property Predict}
Following \citep{altae2017low, wang2021property, zhuang2023graph, wang2024pin}, the FSMPP can be defined as learning to predict molecular properties under limited labeled data per property. Formally, let $\mathcal{T}$ denote the space of property prediction tasks, where each task $T_i \in \mathcal{T}$ corresponds to a binary classification problem associated with a specific molecular property (e.g., toxicity, solubility). Therefore, this is a 2-way K-shot task. Each task $T_i$ consists of a small support set $\mathcal{S}i = {(x_j, y_j)}$,${j=2K}$ of $K$ labeled molecular graphs, and a query set $\mathcal{Q}i = {(x_q, y_q)}$ for evaluation. The goal of FSMPP is to learn a meta-learner $\mathcal{F}\theta$ that can quickly adapt to unseen property prediction tasks using only a few support examples. We need to ensure a separation of properties between the training and testing phases. 
i.e., $\mathcal{P}_{\text{train}} \cap \mathcal{P}_{\text{test}} = \emptyset$.

During meta-training, the model is trained on a set of tasks $\mathcal{T}_{\text{train}}$, where for each episode a task $T_i$ is sampled, its support and query sets are used for inner-loop adaptation and outer-loop meta-update, respectively. During meta-testing, the model is evaluated on a disjoint set of unseen tasks $\mathcal{T}_{\text{test}}$, 
, where for each test task, the model is fine-tuned on $\mathcal{S}_i$ and evaluated on $\mathcal{Q}_i$.

\section{Method}
This section presents the motivation and introduces our M-GLC framework. Figure~\ref{frame} provides an overview of the framework. 
Our method focuses on improving the context graph component. 
In section ~\ref{sec:me1}, we propose a tripartite heterogeneous context graph with \textit{molecule}, \textit{property}, and \textit{motif} nodes.
To address the structural challenges introduced by motif nodes and graph heterogeneity, and to fully exploit their representational power, we further introduce 
the structure-aware edge-weighted aggregation in section ~\ref{sec:me2} 
and the local-focus subgraph in section ~\ref{sec:me3}.
The rest of the pipeline follows prior work \citep{wang2024pin}.

\subsection{Motivation}\label{sec:me0}
Recent work on context‑graph relational embeddings \citep{schimunek2023context, wang2024pin, wang2021property, zhuang2023graph} has advanced few‑shot molecular property learning by supplying property‑conditioned context signals that work as task‑specific priors in meta‑learning, enabling rapid adaptation to unseen properties from few labeled molecules.
However, two challenges remain: 
(i) label sparsity: molecule–property graphs are highly incomplete, with many associations missing even in standard benchmarks \citep{wu2018moleculenet}, which weakens the informativeness of the topology; 
(ii) simple GNNs suffer from representation entanglement: homogeneous aggregation mixes heterogeneous signals and obscures task‑relevant semantics. \citep{zhou2024slotgat}.
Therefore, we propose a tri‑partite motif–molecule–property graph to enrich the context, 
introduce a local‑focus subgraph encoder to concentrate on the most relevant signals, 
and design a structure‑aware aggregation to balance the contributions of motifs, molecules, and properties.
\subsection{Tri-partite Heterogeneous Graph Construction}\label{sec:me1}
\begin{figure*}[t]
\centering
\includegraphics[width=0.85\linewidth,clip,trim=0 270 0 50]{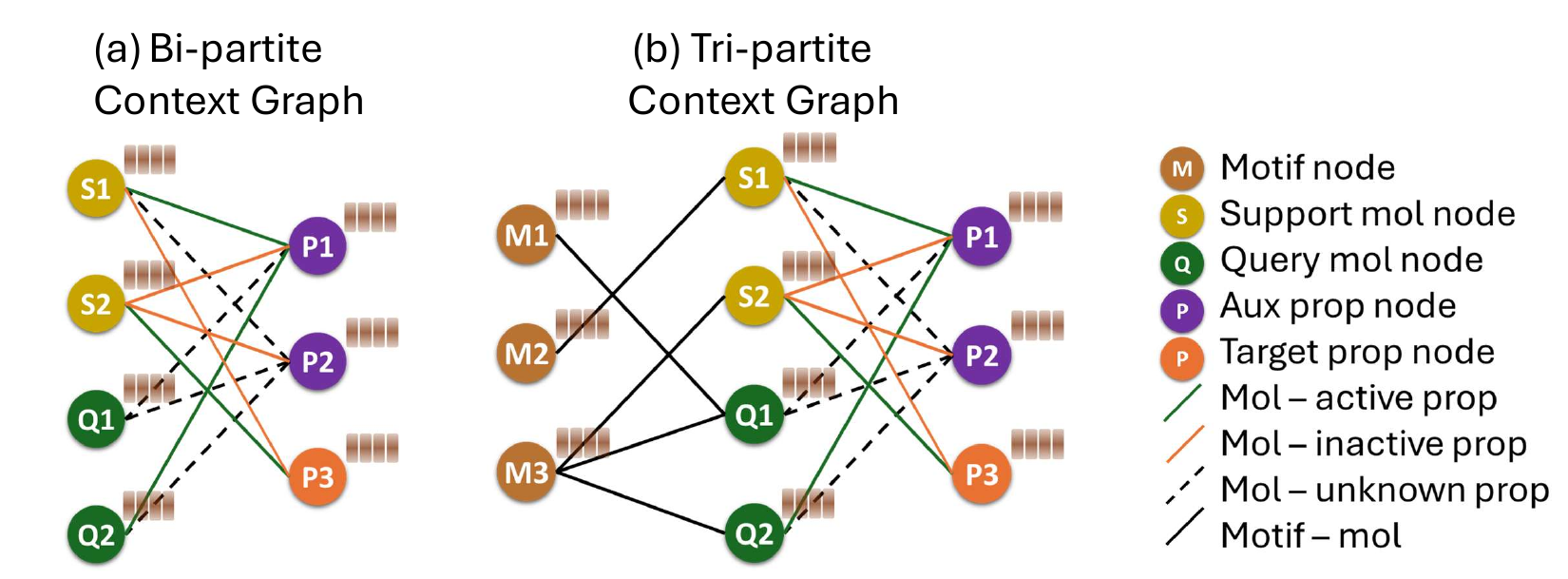}
\Description{}
\caption{ (a) Molecule–property context graph used in previous work \citep{schimunek2023context,wang2024pin,wang2021property,zhuang2023graph}; missing (\textit{unknown}) labels weaken the context information. 
(b) Our tri‑partite context graph introduces domain‑specific motif priors that enhance context informativeness. However, it also increases motif–molecule signal imbalance and heterogeneity, which we mitigate via structure‑aware aggregation and local‑focus subgraph.}
\label{construction}
\end{figure*}

Informative priors are essential for few‑shot learning. Property priors are unreliable when labels are sparse or missing. An example in Fig.~\ref{construction}a, the query molecule $Q1$ has \texttt{unknown} relations to all auxiliary properties, so the bi‑partite graph may fail to provide a usable property prior and adaptation is bottlenecked. We therefore introduce motif‑based structural priors that support molecules via shared substructures, supplying complementary context beyond properties.

\textbf{Motif Dictionary Construction:}
We define a motif \( z_i = (\mathcal{V}_i, \mathcal{E}_i) \) as a subgraph of a molecule \( \mathcal{G} \).  
We follow \citet{jin2020hierarchical} to extract motifs from molecules. In particular, we find all bridge bonds \((u, v)\), where both \( u \) and \( v \) have degrees \( \Delta_u, \Delta_v \geq 2 \), and either \( u \) or \( v \) is part of a ring. 
After detaching all bridge bonds, we obtain motif \( z_i \) as a subgraph of \( \mathcal{G} \). 

\begin{figure}[htbp]
  \centering
  \includegraphics[width=1.0\linewidth]{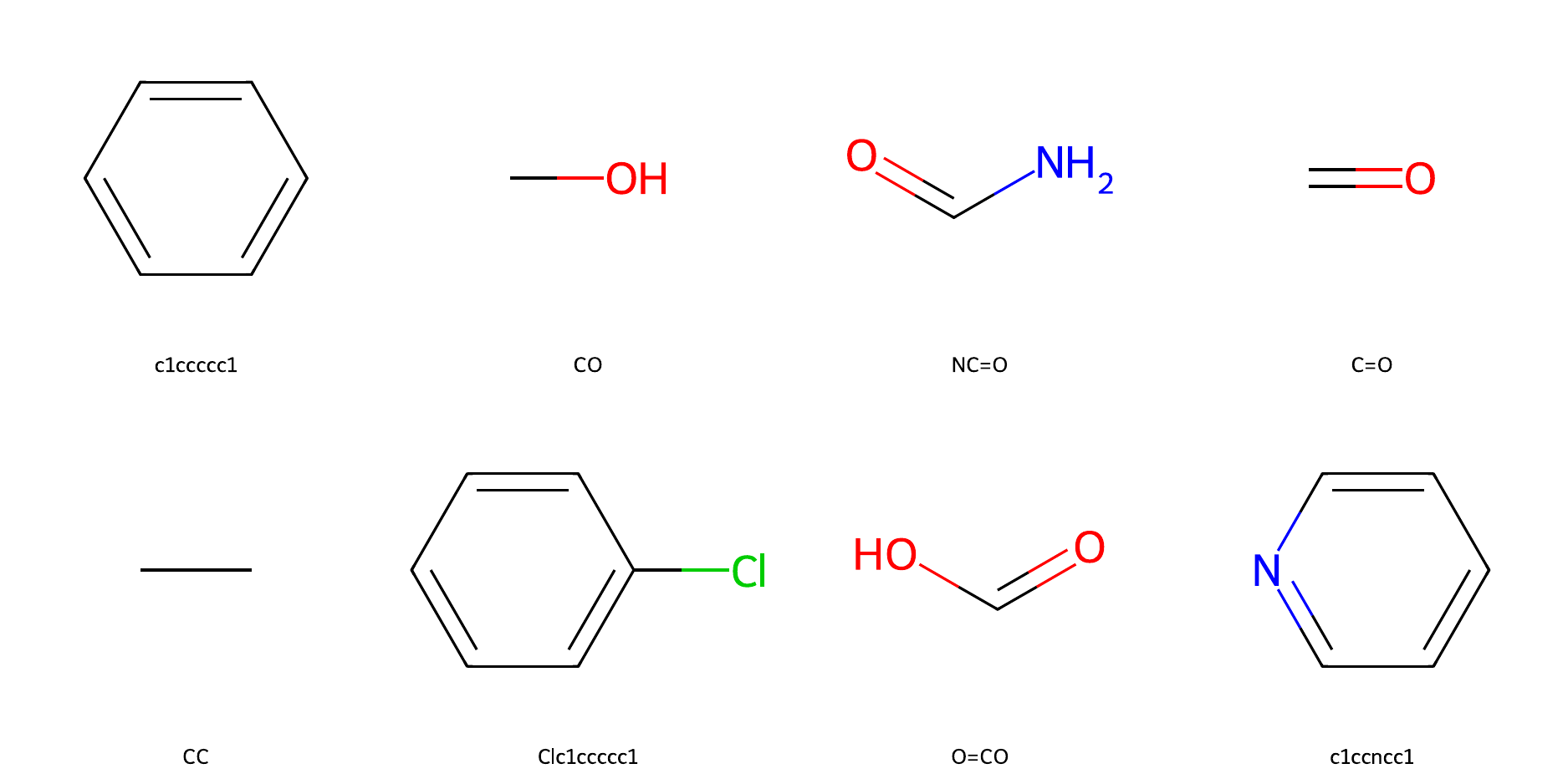} 
  \Description{}
  \caption{The top–8 motifs by frequency in ChEMBL \citep{gaulton2012chembl}, obtained via the bridge extraction method \citep{jin2020hierarchical}. See Appendix~\ref{visualization} for more high-frequency motifs.
    }
  \label{8motif}
\end{figure}

We extract motifs and construct a motif dictionary from the ChEMBL dataset \citep{gaulton2012chembl}. 
We select the top \( K \) motifs with the highest frequencies to form the motif dictionary \( \mathcal{Z} = \{z_1, z_2, \dots, z_K\} \).

\textbf{Tri-partite Graph Node Construction:}
To take advantage of the relationships among motifs, molecules, and molecular properties, we construct a tri-partite heterogeneous graph denoted as \( \mathcal{G} = (\mathcal{V}, \mathcal{E}) \). This graph explicitly augments the connections among molecules. In this context, \( \mathcal{V} \) is the set of nodes, and \( \mathcal{E} \) is the set of edges.
The node set \( \mathcal{V} \) is composed of three subsets: the motif node set \( \mathcal{V}_z \), the molecule node set \( \mathcal{V}_m \), and the property node set \( \mathcal{V}_p \). Thus, we have \( \mathcal{V} = \mathcal{V}_z \cup \mathcal{V}_m \cup \mathcal{V}_p \).

\textbf{Tri-partite Graph Edge Construction:}
As shown in Fig~\ref{construction}b, there are two types of edges in this graph
The first type, \textbf{molecule–property} edges, we use the same strategy from previous work \citep{wang2024pin}. \( \mathcal{E}_{mp} \) connects molecule nodes \( u \in \mathcal{V}_m \) to property nodes \( v \in \mathcal{V}_p \). The edge attribute is the relationship between a molecule and a property defined by the label \( y_{u,v} \in \{1, 0, \texttt{NaN}\} \). For molecules in the support set, we connect them to all property nodes. For molecules in the query set,  we connect them only to auxiliary property nodes.
The second type, \textbf{motif–molecule} edges \( \mathcal{E}_{zm} \), connects each molecule node \( u_i \in \mathcal{V}_m \) to its motifs \( v_j \in \mathcal{V}_z \) from the motif dictionary. We assign these edges a distinct type to differentiate them from those in \( \mathcal{E}_{mp} \). We instantiate the edge set of our tri-partite graph as
$\mathcal{E} \;=\; \mathcal{E}_{mp} \;\cup\; \mathcal{E}_{zm}.$
For both edge types, we embed the edge attributes into a learnable vector space
and initialize all parameters with Xavier initialization \citep{pmlr-v9-glorot10a}.

\begin{figure*}[ht]
  \centering
  \includegraphics[width=0.95\linewidth,clip,trim=0 270 0 0]{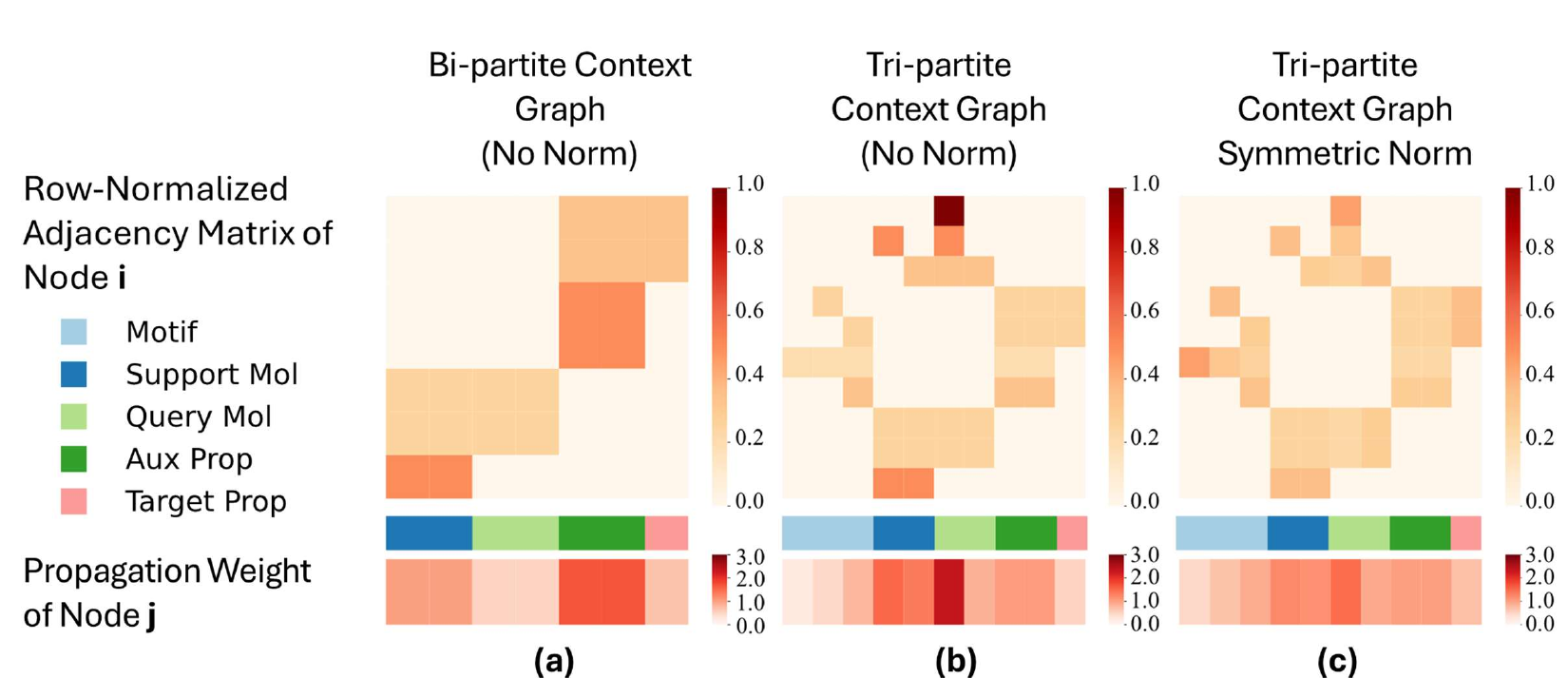} 
  \Description{}
  \caption{
    (a) In the previous context graph, edge weights are the same within each node type and influence message passing equally.
    (b) In our proposed tri-partite context graph, edge weights are uneven. Even among nodes of the same type, such as motifs or molecules, the contributions vary significantly. Some motif nodes have minimal influence (e.g., node 0 in the motif set), while a few molecule nodes dominate the overall propagation weight distribution (e.g., node 5 in the molecule set). Node indices correspond to positions shown in the figure.
    (c) After normalization, the imbalance observed in (b) is reduced, resulting in a more balanced flow of information.
  }
  \label{fig: norm}
\end{figure*}

\subsection{Structure-Aware Edge-Weighted Aggregation}\label{sec:me2}
While extending the context graph to a tri‑partite structure enriches structural priors, it also introduces degree bias on the molecule–motif side.

\textbf{Identical weighted message passing in bipartite context graph.}
In the molecule–property context graphs used by prior work \citep{wang2024pin, zhuang2023graph}, edges are dense and carry categorical labels (\textit{active}, \textit{inactive}, \textit{unknown}). 
Each kind of node has a fixed and identical neighborhood size. With a mean aggregator and unit edge weights, message passing can be reduced to the Row‑Normalized Adjacency matrix $A_{RN}$:
\begin{equation}
A_{RN} \;=\; D^{-1}A,
\end{equation}
Hence, nodes within the same node set aggregate over equally sized neighborhoods. Also shown in Figure~\ref {fig: norm}.a, this also implies column‑wise uniformity: within each node set, all source nodes contribute the same total propagated weight to their neighbors.
This uniformity removes degree bias and makes uniform edge weights a natural choice for the GNN encoder.

\textbf{Identical weighted message passing in tri-partite context graph.}
When the motif-molecule side is introduced, the high-frequency motifs are contained in more molecules, and larger molecules contain more motifs.
As shown in Figure~\ref{fig: norm}b, column‑wise uniformity is broken. Some nodes contribute much more total propagation weight to their neighbors, leading to unbalanced message passing. 
This degree bias in GNNs can degrade representation quality \citep{tang2020investigating}. 

\textbf{Structure-aware edge-weighted message passing in tri-partite context graph.}
We introduce symmetric normalization \citep{kipf2017gcn} to the adjacency matrix, which balances the influence of nodes during message passing in the tri-partite context graph. Also, edge attributes are critical in our model, as they encode auxiliary labels and other task‑relevant signals beyond topology. Accordingly, rather than a plain GCN, we use an MPNN that combines normalized edge weights with edge‑type embeddings. The node signal propagation of the new structure-aware edge-weighted
aggregation can be simplified to the following Row‑Normalized operator:
\begin{equation}
A_{\mathrm{GCN}} = D^{-1/2} A D^{-1/2} 
\end{equation}
\begin{equation}
D'_{ii} \;=\; \sum_{j} \frac{A_{ij}}{\sqrt{d_i\,d_j}},
\qquad d_i = \sum_j A_{ij} 
\end{equation}
\begin{equation}
A_{\mathrm{RN}}' = (D')^{-1} A_{\mathrm{GCN}} 
\end{equation}
The resulting message passing of the context GNN encoder is:
\begin{equation}
w_{ij} = \frac{1}{\sqrt{d_i d_j}}  
\end{equation}
\begin{equation}
\mathbf{h}_i^{(k+1)} = \sum_{j \in \mathcal{N}(i)} w_{ij}\,\bigl(\mathbf{h}_j^{(k)} + \mathbf{e}_{ij}\bigr) + \mathbf{h}_i^{(k)} 
\end{equation}
As illustrated in Fig.~\ref{fig: norm}c, the column‑wise disparity in total propagation weight among nodes is reduced relative to Fig.~\ref{fig: norm}b.

\subsection{Local-Focus Subgraph} \label{sec:me3}
In previous work \citep{wang2024pin, zhuang2023graph}, the context information extract from the context graph can be considered as: 
\begin{equation}\label{contex_emb_org}
\mathbf H = \psi\!\big( \mathcal{G} ), \qquad
\mathbf h_{\text{mol}} = \mathbf H_{i_{\text{mol}}}, \qquad
\mathbf h_{\text{prop}} = \mathbf H_{i_{\text{prop}}}.
\end{equation}

Here,  
$\psi$ is the GNN encoder, $\mathbf H$ is the node feature extracted from the context graph $\mathcal{G} $,
$\mathbf{h}_{\text{prop}}$ is the property context embedding and $\mathbf{h}_{\text{mol}}$ is molecule context embedding. 

While GNN‑based context encoders support multi‑hop aggregation, they face specific challenges in a heterogeneous context graph setting. 
Especially in a tri‑partite heterogeneous graph, node types carry different semantics and should contribute differently to the representation, and a simple GNN context‑graph encoder will lead to representations being entangled with each other in a heterogeneous graph setting. \citep{zhou2024slotgat} 
To reduce such noise and focus on more relevant patterns, we introduce a local-focus subgraph module. 

\begin{figure*}[ht]
\centering
\includegraphics[width=0.95\linewidth,clip,trim=1 330 0 50]{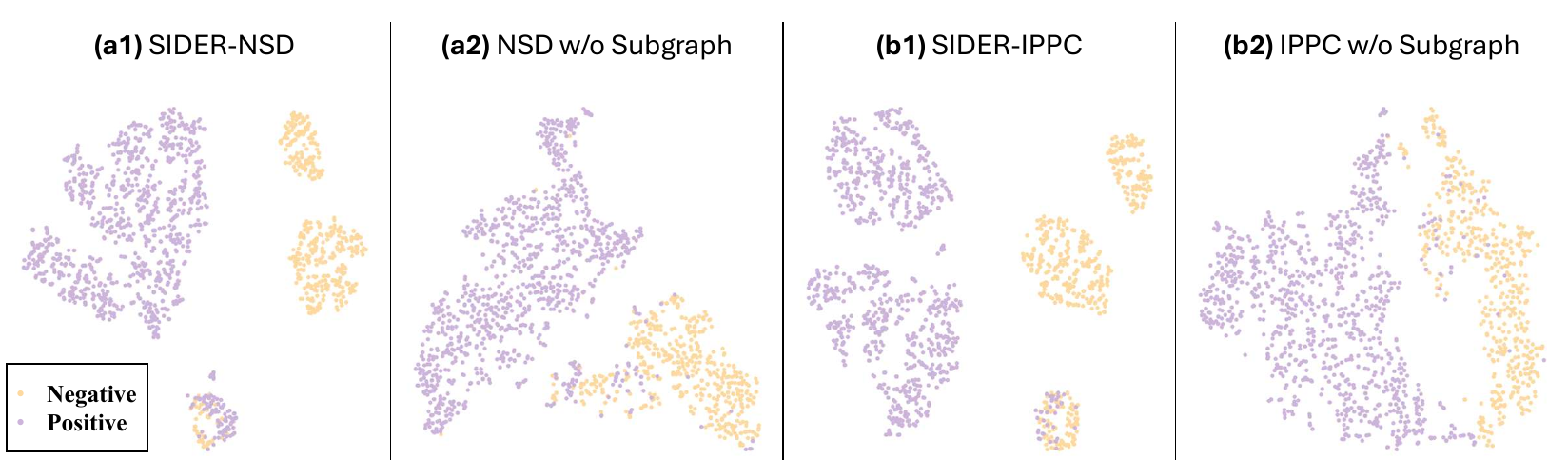} 
\Description{}
\caption{
We project query‑molecule context embeddings at test time to 2D with PCA.
\textbf{(a1)} and \textbf{(b1)} use the local‑focus subgraph; \textbf{(a2)} and \textbf{(b2)} remove it and read out node‑level embeddings.
With the subgraph, embeddings form well‑separated clusters with clear structure within each label, and high‑confidence positives and negatives concentrate into tight groups while potential misclassifications converge into a small, isolated cluster.
Without the subgraph, clusters are diffuse; many positive samples disperse throughout the negative region, indicating stronger noise and single‑node bias.
}
\label{fig: sub_visual}
\end{figure*}

\textbf{Local-Focus Subgraph Definition:}
The local subgraph is defined as the 1-hop neighborhood around the target node, ensuring that only the most relevant nearby nodes are included.

Formally, given a context graph $\mathcal{G} = (\mathcal{V}, \mathcal{E})$ and a target node $v_t \in \mathcal{V}$, we define its local-focus subgraph as:
\begin{equation}
\mathcal{G}_{\text{sub}}^{(t)} = (\mathcal{V}_{\text{sub}}^{(t)}, \mathcal{E}_{\text{sub}}^{(t)}),
\end{equation}
where $\mathcal{V}_{\text{sub}}^{(t)} = \{ v_t \} \cup \mathcal{N}(v_t)$ includes the target node and its one hop neighbors, and $\mathcal{N}(v_t)$ denotes the set of nodes directly connected to $v_t$. The edge set $\mathcal{E}_{\text{sub}}^{(t)}$ contains all edges among nodes in $\mathcal{V}_{\text{sub}}^{(t)}$. A GNN encoder is applied to this subgraph to further encode the local representation $\mathbf{h}_{\text{sub}}^{(t)}$.

\textbf{Local‑Focus Subgraph Encoder:} 
We adopt local subgraphs with dedicated GNN encoders to obtain cleaner representations.
For a molecule‑centric subgraph \(i\), its local‑focus subgraph includes the nodes
\begin{equation} \label{input_mol}
\mathcal{V}_{\text{sub}}^{(i)} = \{\, i \,\}\cup \mathcal N_z(i)\cup \mathcal N_p(i),
\end{equation}
And we denote the subgraph as
\begin{equation}
\mathcal{G}_{\text{sub}}^{(i)} = \big(\mathcal{V}_{\text{sub}}^{(i)},\, \mathcal{E}_{\text{sub}}^{(i)}\big).
\end{equation}
The context embedding is computed by a dedicated molecule encoder:
\begin{equation}\label{contex_emb_mol}
\mathbf h_{\text{mol}}(i) = \sigma\!\left( \phi_{\text{mol}}\!\big(\mathcal{G}_{\text{sub}}^{(i)}\big) \right)
\end{equation}
Here, \(\sigma\) is an aggregation function and \(\phi_{\text{mol}}\) is a dedicated GNN encoder that aggregates interactions among the molecule \(i\), its motif neighbors \(\mathcal N_z(i)\), and property neighbors \(\mathcal N_p(i)\).
For a property‑centric subgraph \(j\), its nodes include
\begin{equation} \label{input_prop}
\mathcal{V}_{\text{sub}}^{(j)} = \{\, j \,\}\cup \mathcal N_m(j),
\end{equation}
with the subgraph
\begin{equation}
\mathcal{G}_{\text{sub}}^{(j)} = \big(\mathcal{V}_{\text{sub}}^{(j)},\, \mathcal{E}_{\text{sub}}^{(j)}\big)
\end{equation}
and the embedding is
\begin{equation}\label{contex_emb_prop}
\mathbf h_{\text{prop}}(j) = \sigma\!\left( \phi_{\text{prop}}\!\big(\mathcal{G}_{\text{sub}}^{(j)}\big) \right).
\end{equation}

Compared with the input to $\phi_{\text{mol}}$ in Eq.~\ref{input_mol}, the input to $\phi_{\text{prop}}$ in Eq.~\ref{input_prop} is simpler: $\phi_{\text{prop}}$ only needs to model interactions between the query property $j$ and its molecule neighbors $\mathcal N_m(j)$. 
Moreover, Eqs.~\ref{contex_emb_mol}–\ref{contex_emb_prop} show that our local‑focus readout operates on the entire subgraph $\mathcal G_{\text{sub}}^{(\cdot)}$, in contrast to the previous node‑level readout in Eq.~\ref{contex_emb_org}.

Therefore, the local‑focus subgraph provides two advantages: (1) Dedicated GNN encoders respect the distinct roles of molecule and property nodes and extract specific context features. (2) Aggregating over the entire local subgraph produces cleaner and more stable embeddings, as evidenced in Fig.~\ref{fig: sub_visual}.

\section{Experimental Result}
We evaluate M-GLC by comparing its performance with previous SOTA FSMPP methods using benchmark datasets.
\subsection{Experiment Settings}
\textbf{Baselines}  
To ensure a comprehensive evaluation, we consider two categories of baseline methods. The first includes models trained from scratch without pertaining: Siamese Network~\citep{koch2015siamese}, ProtoNet~\citep{snell2017prototypical}, MAML~\citep{finn2017model}, TPN~\citep{liu2018learning}, EGNN~\citep{kim2019edge}, and IterRefLSTM~\citep{altae2017low}. The second category consists of methods that leverage pretrained molecular encoders, including Pre-GNN~\citep{hu2019strategies}, Meta-MGNN~\citep{guo2021few}, PAR~\citep{wang2021property}, GS-Meta~\citep{zhuang2023graph}, and Pin-Tuning \citep{wang2024pin}. 

\begin{table*}[t!]
\centering
\caption{Few-shot experimental results on benchmark datasets. ROC-AUC scores (\%) are reported. The top block includes models trained from scratch; the bottom block includes methods using pre-trained molecular encoders. Best results are in bold. Numbers in parentheses denote standard deviation. 'Boost' indicates relative improvement over baseline (in \%).}
\label{tab:mainresult_fewshot}
{\footnotesize
\begin{minipage}{\textwidth}
\centering
\textbf{(a) 10-shot Results}\\[0.5ex]
\begin{tabularx}{\linewidth}{l|Y|Y|Y|Y|Y}
\toprule
\textbf{Method} & \textbf{Tox21} & \textbf{SIDER} & \textbf{MUV} & \textbf{ToxCast} & \textbf{PCBA} \\
\midrule
ProtoNet & 74.98(0.32) & 64.54(0.89) & 65.88(4.11) & 68.87(0.43) & 64.93(1.94) \\
MAML & 80.21(0.24) & 70.43(0.76) & 63.90(2.28) & 68.30(0.59) & 66.22(1.31) \\
TPN & 76.05(0.24) & 67.84(0.95) & 65.22(5.82) & 69.47(0.71) & 67.61(0.33) \\
EGNN & 81.21(0.16) & 72.87(0.73) & 65.20(2.08) & 74.02(1.11) & 69.92(1.85) \\
IterRefLSTM & 81.10(0.17) & 69.63(0.31) & 49.56(5.12) & - & - \\
\midrule
Pre-GNN & 82.14(0.08) & 73.96(0.08) & 67.14(1.58) & 75.31(0.95) & 76.79(0.45) \\
Meta-MGNN & 82.97(0.10) & 75.43(0.21) & 68.99(1.84) & 76.27(0.56) & 72.58(0.34) \\
PAR & 84.93(0.11) & 78.08(0.16) & 69.96(1.37) & 79.41(0.08) & 73.71(0.61) \\
GS-Meta & 86.67(0.41) & 84.36(0.54) & 66.08(1.25) & 83.81(0.16) & 79.40(0.43) \\
Pin-Tuning & 91.56(2.57) & 93.41(3.52) & 73.33(2.00) & 84.94(1.09) & 81.26(0.46) \\
M-GLC (Ours) & \textbf{98.17(0.66)} & \textbf{98.37(0.83)} & \textbf{79.31(1.96)} & \textbf{90.24(1.51)} & \textbf{84.80(1.30)} \\
\midrule
Boost & \textbf{7.22} & \textbf{5.31} & \textbf{8.15} & \textbf{6.24} & \textbf{4.36} \\
\bottomrule
\end{tabularx}
\end{minipage}
\vspace{1em}
\begin{minipage}{\textwidth}
\centering
\textbf{(b) 5-shot Results}\\[0.5ex]
\begin{tabularx}{\linewidth}{l|Y|Y|Y|Y|Y}
\toprule
\textbf{Method} & \textbf{Tox21} & \textbf{SIDER} & \textbf{MUV} & \textbf{ToxCast} & \textbf{PCBA} \\
\midrule
ProtoNet & 72.78(3.93) & 64.09(2.37) & 64.86(2.31) & 66.26(1.49) & 62.29(2.12) \\
MAML & 69.17(1.34) & 60.92(0.65) & 63.00(0.61) & 67.56(1.53) & 65.25(0.75) \\
TPN & 75.45(0.95) & 66.52(1.28) & 65.13(0.23) & 66.04(1.14) & 63.66(1.64) \\
EGNN & 76.80(2.62) & 60.61(1.06) & 63.46(2.58) & 67.13(0.50) & 67.71(3.67) \\
IterRefLSTM & - & - & - & - & - \\
\midrule
Pre-GNN & 82.04(0.30) & 76.76(0.53) & 70.23(1.40) & 74.43(0.47) & 75.27(0.49) \\
Meta-MGNN & 76.12(0.23) & 66.60(0.38) & 64.07(0.56) & 75.26(0.43) & 72.51(0.52) \\
PAR & 83.95(0.15) & 77.70(0.34) & 68.08(2.42) & 76.89(0.32) & 72.79(0.98) \\
GS-Meta & 86.43(0.02) & 84.57(0.01) & 64.50(0.20) & 82.65(0.35) & 77.47(0.29) \\
Pin-Tuning & 90.95(2.33) & 92.02(3.01) & 70.71(1.42) & 83.71(0.93) & 79.23(0.52) \\
M-GLC (Ours) & \textbf{98.39(0.85)} & \textbf{98.89(0.57)} & \textbf{79.09(2.97)} & \textbf{90.35(1.62)} & \textbf{85.13(2.20)} \\
\midrule
Boost & \textbf{8.18} & \textbf{7.46} & \textbf{11.85} & \textbf{7.93} & \textbf{7.45} \\
\bottomrule
\end{tabularx}
\end{minipage}
}
\end{table*}

\textbf{Datasets}  
Following prior works~\citep{wang2021property, zhuang2023graph, wang2024pin}, we evaluate on five widely-used datasets for few-shot molecular property prediction, all sourced from MoleculeNet: Tox21, SIDER, MUV, ToxCast, and PCBA. The dataset statistics can be found in the Table.~\ref{tab:dataset_statistics}

\begin{table}[t]
\centering
\caption{Dataset statistics for molecular property prediction benchmarks.}
\label{tab:dataset_statistics}
{\footnotesize
\begin{tabularx}{\linewidth}{l|YYYYY}
\toprule
\textbf{Dataset} & \textbf{Tox21} & \textbf{SIDER} & \textbf{MUV} & \textbf{ToxCast} & \textbf{PCBA} \\
\midrule
\#Compound        & 7,831  & 1,427   & 93,127   & 8,575   & 437,929 \\
\#Property        & 12     & 27      & 17       & 617     & 128     \\
\#Train Property  & 9      & 21      & 12       & 451     & 118     \\
\#Test Property   & 3      & 6       & 5        & 158     & 10      \\
\%Positive Label  & 6.24\% & 56.76\% & 0.31\%   & 12.60\% & 0.84\%  \\
\%Negative Label  & 76.71\%& 43.24\% & 15.76\%  & 72.43\% & 59.84\% \\
\%Unknown Label   & 17.05\%& 0\%     & 84.21\%  & 14.97\% & 39.32\% \\
\bottomrule
\end{tabularx}
}
\end{table}

\textbf{Evaluation Metrics}  
Following prior works~\citep{finn2017model, wang2021property, wang2024pin, zhuang2023graph}, we report the ROC-AUC score computed on the query set of each meta-testing task, which reflects the model’s ability to generalize under the few-shot setting. 
Following previous works\citep{wang2024pin, zhuang2023graph}, we report the average performance across 10 random seeds. 
Following~\citet{wang2024pin}, we evaluate our method under both 5-shot and 10-shot configurations.

\subsection{Main Result}
We compare our method with both from-scratch and pre-trained baselines. Results are shown in Table~\ref{tab:mainresult_fewshot}. Across five benchmarks, our method achieves higher ROC-AUC scores than all baselines in both the 5-shot and 10-shot settings.

\captionsetup[subfigure]{justification=centering}
\begin{figure*}[t!]
  \Description{}
  \centering
  \begin{subfigure}[b]{0.24\textwidth}
    \caption*{SIDER IPPC\\10‑shot M‑GLC}
    \includegraphics[width=\textwidth]{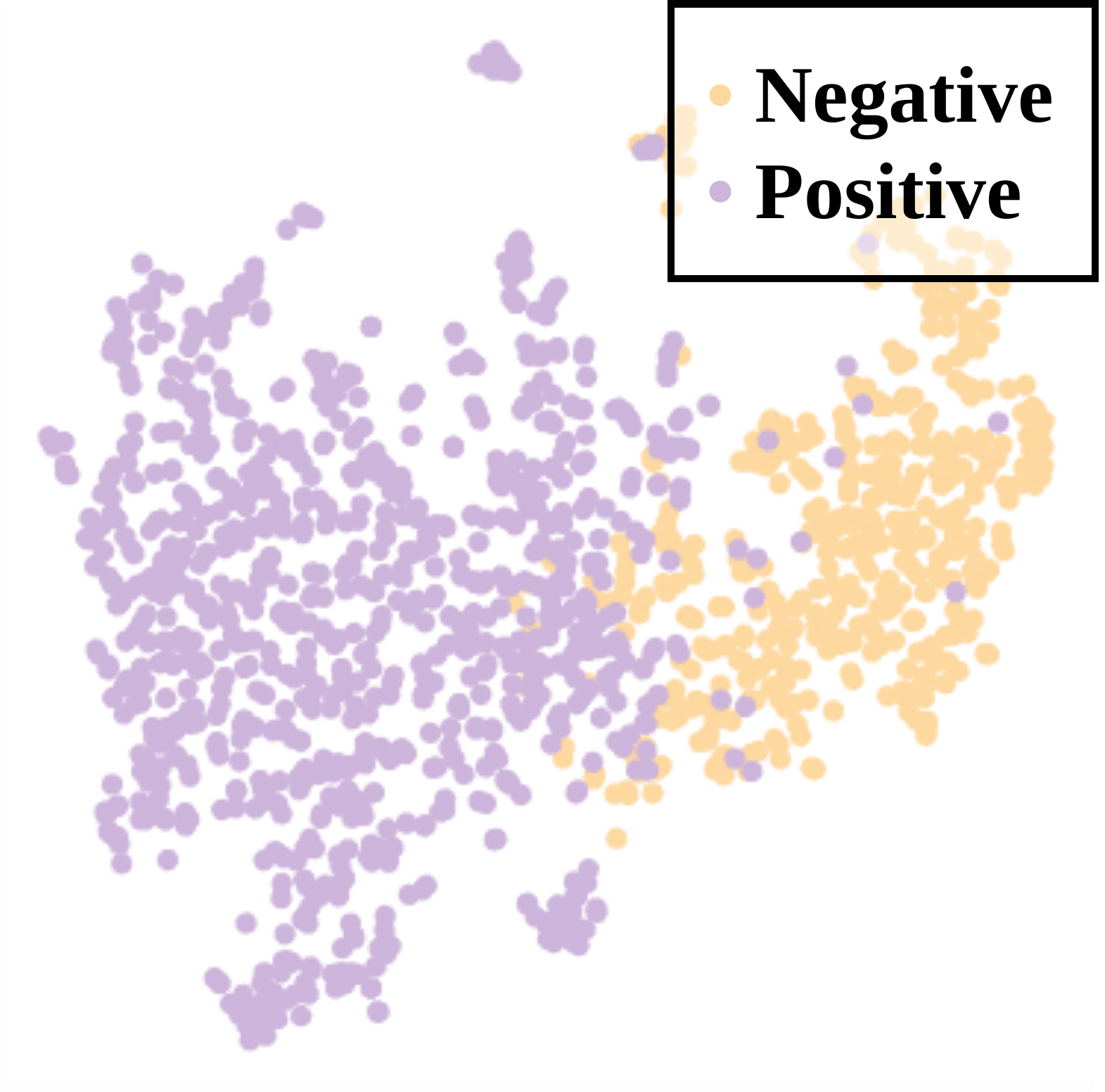}
  \end{subfigure}\hfill
  \begin{subfigure}[b]{0.24\textwidth}
    \caption*{SIDER IPPC\\10‑shot Pin‑Tuning}
    \includegraphics[width=\textwidth]{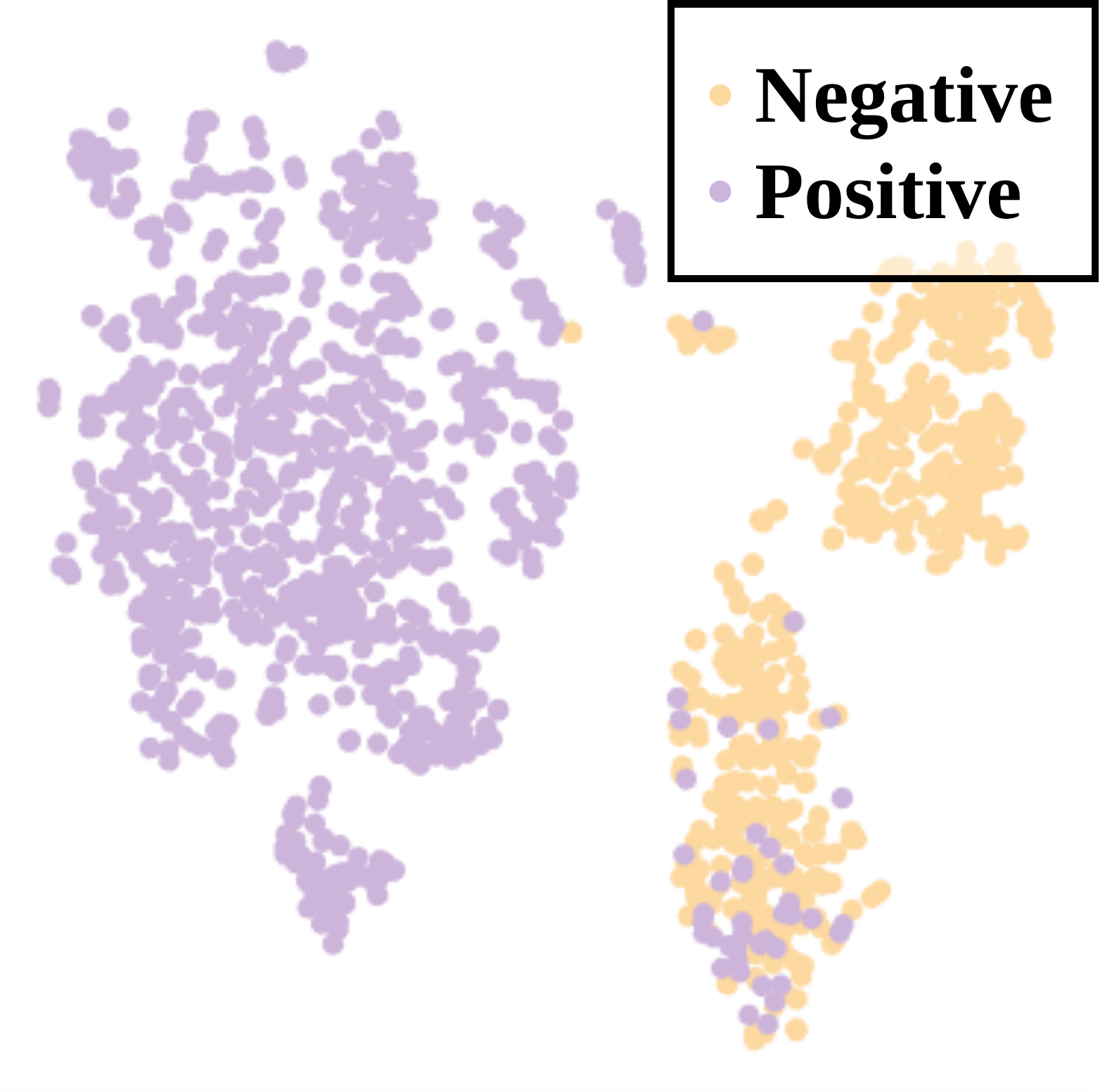}
  \end{subfigure}\hfill
  \begin{subfigure}[b]{0.24\textwidth}
    \caption*{SIDER RUD\\10‑shot M‑GLC}
    \includegraphics[width=\textwidth]{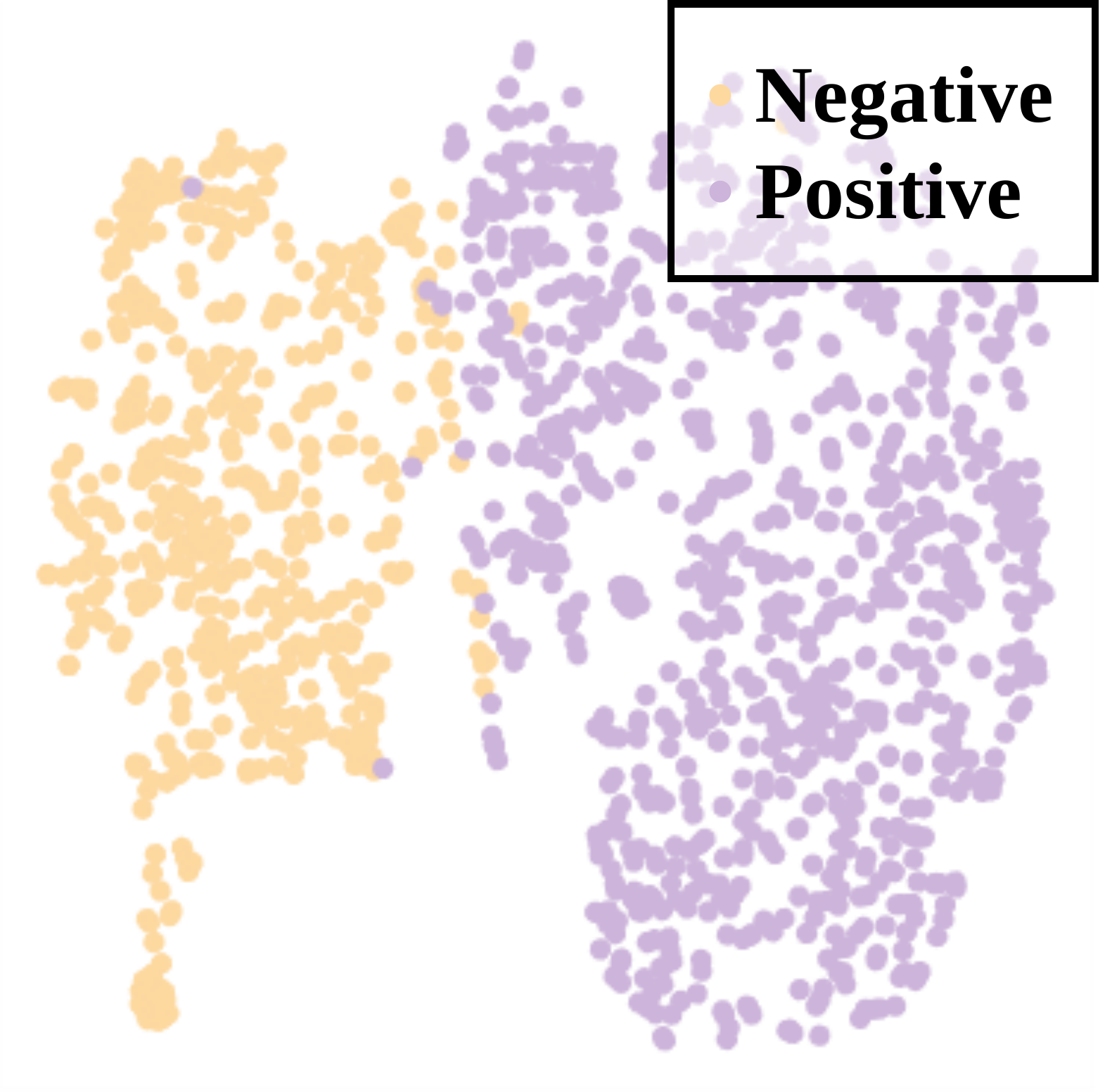}
  \end{subfigure}\hfill
  \begin{subfigure}[b]{0.24\textwidth}
    \caption*{SIDER RUD\\10‑shot Pin‑Tuning}
    \includegraphics[width=\textwidth]{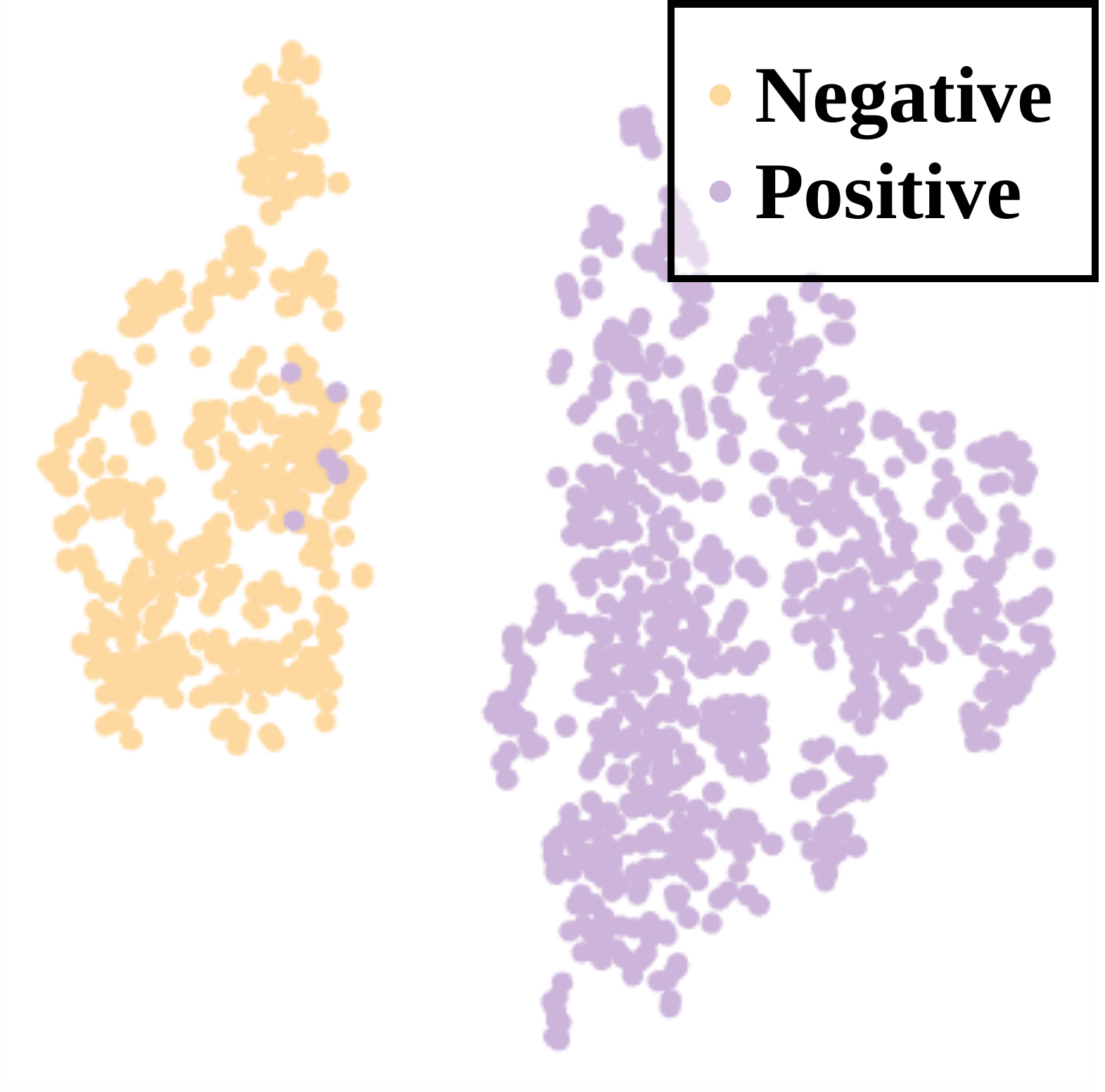}
  \end{subfigure}

  \vspace{1ex}

  \begin{subfigure}[b]{0.24\textwidth}
    \caption*{MUV‑852 FXIIa\\10‑shot M‑GLC}
    \includegraphics[width=\textwidth]{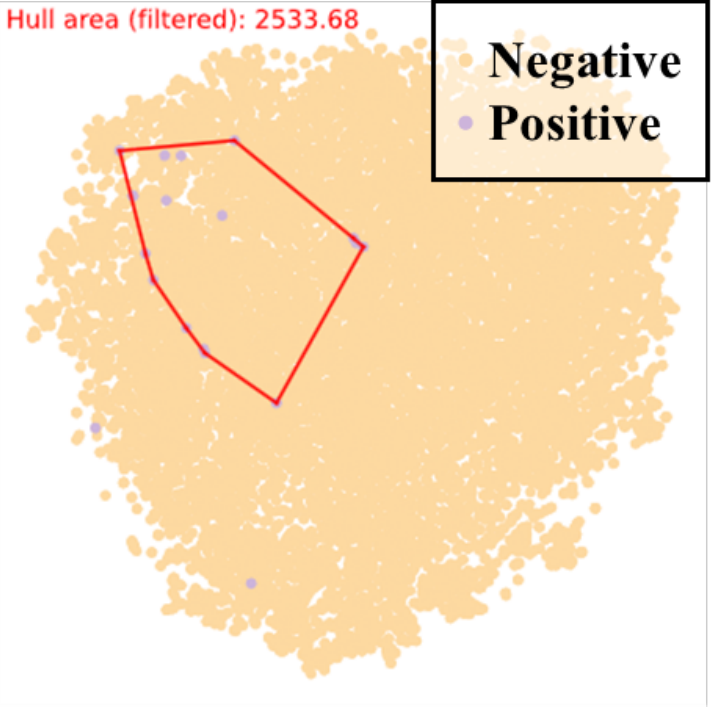}
  \end{subfigure}\hfill
  \begin{subfigure}[b]{0.24\textwidth}
    \caption*{MUV‑852 FXIIa\\10‑shot Pin‑Tuning}
    \includegraphics[width=\textwidth]{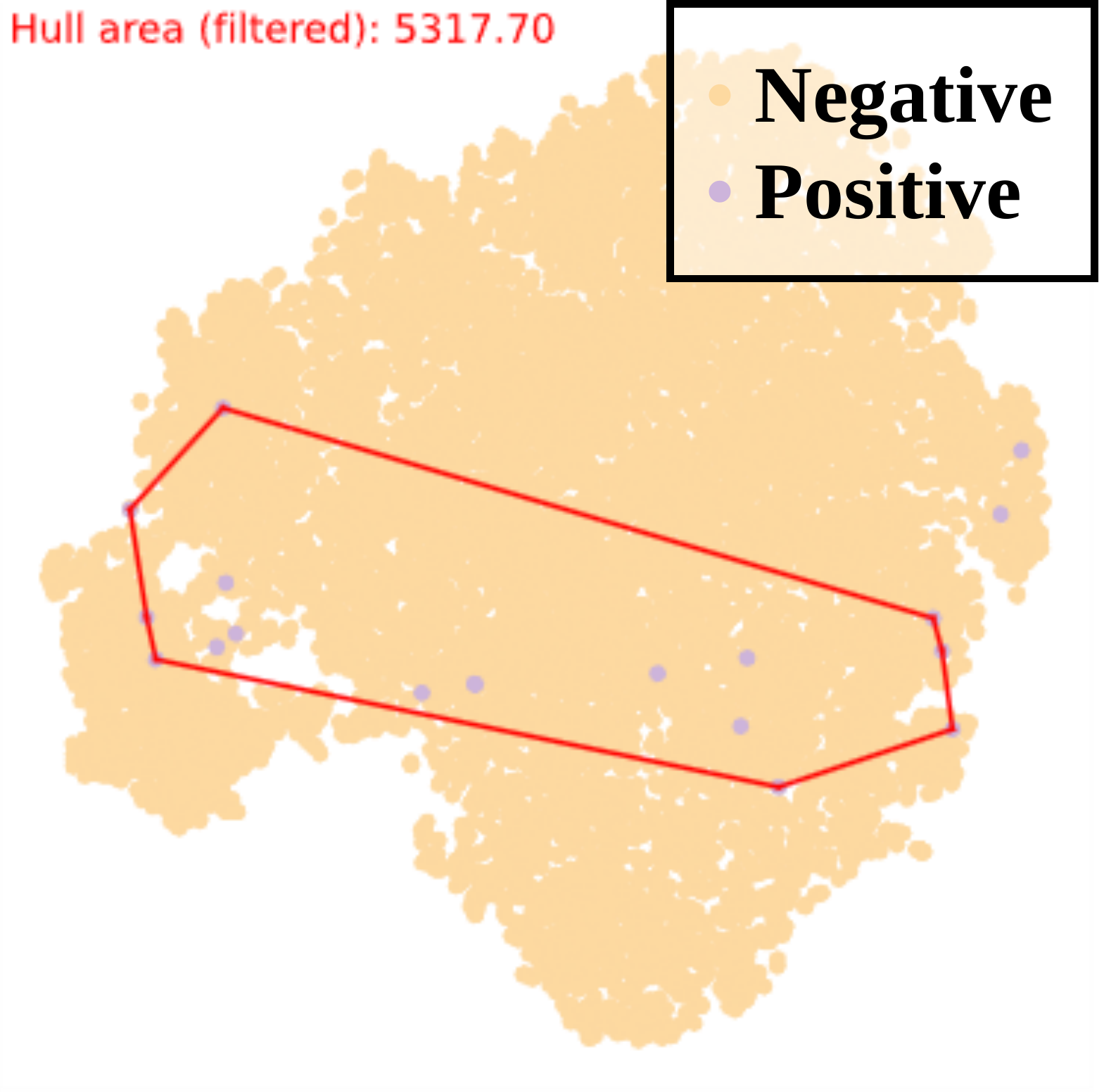}
  \end{subfigure}\hfill
  \begin{subfigure}[b]{0.24\textwidth}
    \caption*{MUV‑859 M1 rec\\10‑shot M‑GLC}
    \includegraphics[width=\textwidth]{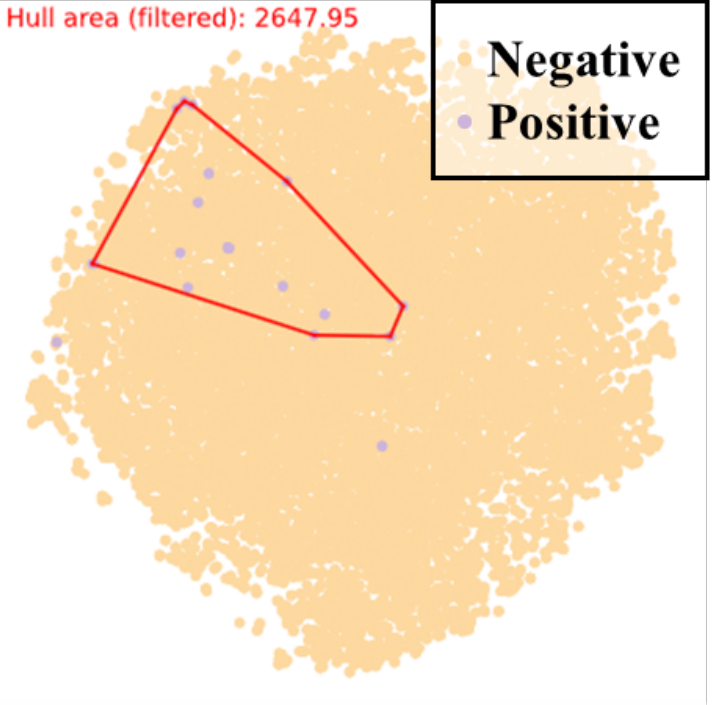}
  \end{subfigure}\hfill
  \begin{subfigure}[b]{0.24\textwidth}
    \caption*{MUV‑859 M1 rec\\10‑shot Pin‑Tuning}
    \includegraphics[width=\textwidth]{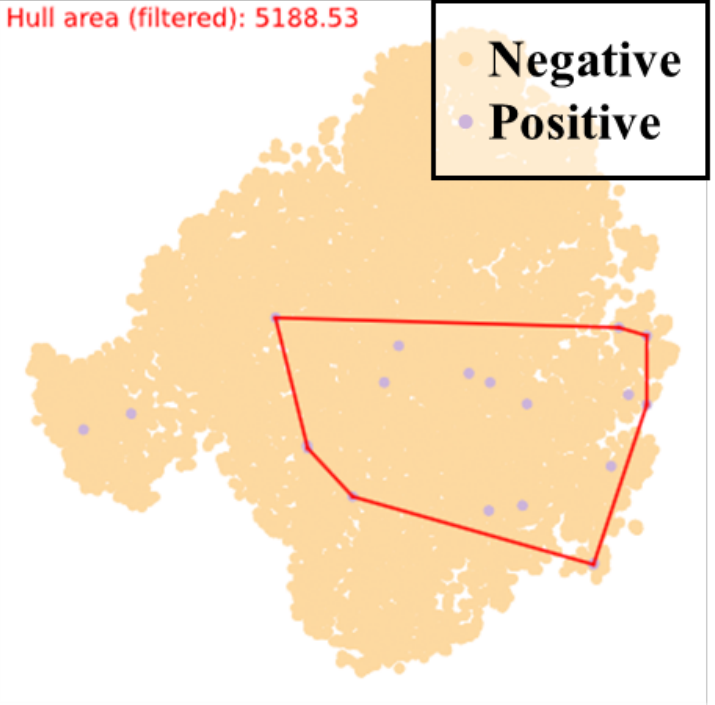}
  \end{subfigure}
\caption{2D projections of positive and negative samples from four tasks: two SIDER tasks (IPPC and RUD) and two MUV tasks (852 FXIIa and 859 M1 rec). To highlight positive-sample spread of Muv tasks, red convex hulls enclose the central 90\% based on distance to the class centroid, removing a few outliers to avoid overestimating dispersion.}
\label{fig:combined-case}
\end{figure*}

On \textbf{Tox21}, \textbf{SIDER}, \textbf{ToxCast}, and \textbf{PCBA}, we observe consistent improvements ranging from 4.36\% to 8.18\%. For Tox21 and SIDER, the final scores exceed 98\%, and the standard deviations are small, showing that the method performs reliably across different runs. 
The performance of our method on the nine ToxCast sub-datasets is reported in Tables~\ref{tab:toxcast-5shot} and~\ref{tab:toxcast-10shot}.

On \textbf{MUV}, the improvements are larger: 8.15\% in the 10-shot setting and 11.85\% in the 5-shot setting. This is likely because MUV provides weaker auxiliary property signals compared to other datasets. In this case, the motif-level structural information plays a more important role, helping the model fill in missing context and improving prediction accuracy. 

While the improvements on MUV and PCBA are significant, the variance in the 5-shot setting is also higher, as shown by larger standard deviations. A possible reason is that the model updates motif-related weights using only a few support samples. If the support set does not include representative motifs for the query set, the learned parameters may be biased, especially in large and diverse datasets like PCBA. This can lead to less stable performance.

Even so, our method performs better than all baselines. In all cases, the lowest scores of our model are close to, or higher than, the best scores from previous methods.

\subsection{Ablation Study}
\begin{table}[t]
\centering
\caption{Ablation study on the SIDER and MUV datasets. We report ROC-AUC (\%) scores under the 10-shot and 5-shot settings. Each row corresponds to the removal of one or more components from the M-GLC framework.}
\label{tab:ablation}
\small
\begin{tabularx}{\columnwidth}{c|c|c|YY|YY}
\toprule
Tri-partite  & Context Graph & Subgraph  & 
\multicolumn{2}{c|}{\textbf{MUV}} & 
\multicolumn{2}{c}{\textbf{SIDER}} \\
Graph & Weight Norm & Embedding & 10-shot & 5-shot & 10-shot & 5-shot \\
\midrule
\ding{51} & \ding{51} & \ding{51} & 79.31 & 79.09 & 98.37 & 98.89 \\
          &           & \ding{51} & 76.95 & 73.89 & 92.54 & 92.58 \\
\ding{51} &           & \ding{51} & 75.58 & 74.47 & 92.06 & 93.02 \\
\ding{51} & \ding{51} &           & 74.94 & 74.75 & 83.40 & 89.89 \\
\bottomrule
\end{tabularx}
\end{table}

We conduct ablation studies on two representative datasets, Sider and MUV, which reflect different data characteristics. The Sider dataset is clean and balanced, with 56.76\% positive samples, 43.24\% negative samples, and no missing labels. It serves as a reliable benchmark to evaluate the contribution of the proposed context graph component to FSMPP. In contrast, the MUV dataset is highly imbalanced and sparse, containing only 0.31\% positive samples and 15.76\% negative samples, and 84.21\% missing labels. This represents a more challenging setting in terms of class imbalance and label sparsity.
By evaluating these two datasets, we can examine the impact of each model component under different task difficulty levels, from clean and balanced to noisy and incomplete. Detailed results are summarized in Table \ref{tab:ablation}.

The proposed M-GLC framework consists of three core components: a tripartite context graph, structure-aware edge weight normalization, and a local focus subgraph module. We conducted a series of ablation experiments, gradually removing each component to evaluate its respective contributions.
(1) Removing the \textbf{motif part} and \textbf{structure-aware edge weight normalization} of the three-part context graph simultaneously resulted in a significant drop in performance, confirming the importance of the motif context information and normalization strategy to the overall model.
(2) Removing only \textbf{structure-aware edge weight normalization} resulted in a larger performance drop than (1). Without normalization, high-degree nodes dominate the message passing, introducing noise and weakening the effectiveness of context information.
(3) Removing \textbf{local focus subgraph} resulted in a significant drop in performance, even lower than the baseline method in some cases. This suggests that although the global context graph provides rich structural signals, the lack of local views leads to the mixing of irrelevant information, making the learned embedding less consistent with the target molecule-attribute pair.

Altogether, these results indicate that the three components are closely integrated and mutually reinforcing. Removing any one of them leads to a significant performance drop, highlighting the necessity of a complete design.

\subsection{Case Study}
To better understand how our model behaves under different data conditions, we compare two representative datasets: SIDER and MUV. SIDER has relatively balanced numbers of positive and negative samples, which allows us to observe how the model separates classes when both are well represented. In contrast, MUV is highly imbalanced, with positive samples often several hundred times fewer than negatives. This makes it a strong test case for evaluating how the model responds to rare signals. These two datasets cover complementary scenarios and help us assess the model’s generalization under both balanced and imbalanced settings.

As shown in Fig.~\ref{fig:combined-case}, for SIDER tasks, our method shows clear differences in error patterns compared to the baseline. Most errors made by our model occur near the decision boundary, suggesting that it makes cautious predictions when the input is hard to classify. The baseline, in contrast, produces overconfident errors, with positive samples misclassified deep inside the negative region. This often leads to incorrect predictions made with high certainty. Our use of motif nodes helps reduce such errors by connecting molecules that share common substructures. This promotes better feature alignment and supports more reliable decisions, especially when data is limited. On MUV tasks, where positive samples are extremely scarce, our method produces a more structured distribution of positives in the feature space. While they are not perfectly clustered, the positive samples tend to concentrate in a specific region, showing greater internal consistency. In contrast, the baseline (Pin-Tuning) yields a more scattered distribution, with positive samples spread throughout the negative background and lacking clear grouping. This makes it harder to distinguish active compounds from inactive ones and suggests that the baseline struggles to capture shared features among rare positives. These results indicate that our model helps form more coherent and transferable representations, even under severe class imbalance.

\section{Conclusion}
In this work, we propose M-GLC for few-shot molecular property prediction that uses motif information to expand the context graph, while local subgraphs help the model focus on relevant patterns for each prediction. Our method achieves consistent improvements over previous approaches across multiple benchmarks. In addition, the model produces more boundary-aware predictions, reducing overconfident errors. These results show that incorporating structural inductive bias is effective for improving generalization in few-shot settings, and that motif-aware context is a useful direction for molecular representation learning.

\newpage
\bibliographystyle{ACM-Reference-Format}
\bibliography{reference}


\begin{thebibliography}{44}


\ifx \showCODEN    \undefined \def \showCODEN     #1{\unskip}     \fi
\ifx \showISBNx    \undefined \def \showISBNx     #1{\unskip}     \fi
\ifx \showISBNxiii \undefined \def \showISBNxiii  #1{\unskip}     \fi
\ifx \showISSN     \undefined \def \showISSN      #1{\unskip}     \fi
\ifx \showLCCN     \undefined \def \showLCCN      #1{\unskip}     \fi
\ifx \shownote     \undefined \def \shownote      #1{#1}          \fi
\ifx \showarticletitle \undefined \def \showarticletitle #1{#1}   \fi
\ifx \showURL      \undefined \def \showURL       {\relax}        \fi
\providecommand\bibfield[2]{#2}
\providecommand\bibinfo[2]{#2}
\providecommand\natexlab[1]{#1}
\providecommand\showeprint[2][]{arXiv:#2}

\bibitem[Alsentzer et~al\mbox{.}(2020)]%
        {alsentzer2020subgraph}
\bibfield{author}{\bibinfo{person}{Emily Alsentzer}, \bibinfo{person}{Samuel Finlayson}, \bibinfo{person}{Michelle Li}, {and} \bibinfo{person}{Marinka Zitnik}.} \bibinfo{year}{2020}\natexlab{}.
\newblock \showarticletitle{Subgraph neural networks}.
\newblock \bibinfo{journal}{\emph{Advances in Neural Information Processing Systems}}  \bibinfo{volume}{33} (\bibinfo{year}{2020}), \bibinfo{pages}{8017--8029}.
\newblock


\bibitem[Altae-Tran et~al\mbox{.}(2017)]%
        {altae2017low}
\bibfield{author}{\bibinfo{person}{Han Altae-Tran}, \bibinfo{person}{Bharath Ramsundar}, \bibinfo{person}{Aneesh~S Pappu}, {and} \bibinfo{person}{Vijay Pande}.} \bibinfo{year}{2017}\natexlab{}.
\newblock \showarticletitle{Low data drug discovery with one-shot learning}.
\newblock \bibinfo{journal}{\emph{ACS central science}} \bibinfo{volume}{3}, \bibinfo{number}{4} (\bibinfo{year}{2017}), \bibinfo{pages}{283--293}.
\newblock


\bibitem[Chen et~al\mbox{.}(2023b)]%
        {dlmpp1}
\bibfield{author}{\bibinfo{person}{Dingshuo Chen}, \bibinfo{person}{Yanqiao Zhu}, \bibinfo{person}{Jieyu Zhang}, \bibinfo{person}{Yuanqi Du}, \bibinfo{person}{Zhixun Li}, \bibinfo{person}{Qiang Liu}, \bibinfo{person}{Shu Wu}, {and} \bibinfo{person}{Liang Wang}.} \bibinfo{year}{2023}\natexlab{b}.
\newblock \showarticletitle{Uncovering neural scaling laws in molecular representation learning}.
\newblock \bibinfo{journal}{\emph{Advances in Neural Information Processing Systems}}  \bibinfo{volume}{36} (\bibinfo{year}{2023}), \bibinfo{pages}{1452--1475}.
\newblock


\bibitem[Chen et~al\mbox{.}(2023a)]%
        {chen2022meta}
\bibfield{author}{\bibinfo{person}{Wenlin Chen}, \bibinfo{person}{Austin Tripp}, {and} \bibinfo{person}{Jos{\'{e}}~Miguel Hern{\'{a}}ndez{-}Lobato}.} \bibinfo{year}{2023}\natexlab{a}.
\newblock \showarticletitle{Meta-learning Adaptive Deep Kernel Gaussian Processes for Molecular Property Prediction}. In \bibinfo{booktitle}{\emph{{ICLR}}}. \bibinfo{publisher}{OpenReview.net}.
\newblock


\bibitem[Fang et~al\mbox{.}(2022)]%
        {fang2022geometry}
\bibfield{author}{\bibinfo{person}{Xiaomin Fang}, \bibinfo{person}{Lihang Liu}, \bibinfo{person}{Jieqiong Lei}, \bibinfo{person}{Donglong He}, \bibinfo{person}{Shanzhuo Zhang}, \bibinfo{person}{Jingbo Zhou}, \bibinfo{person}{Fan Wang}, \bibinfo{person}{Hua Wu}, {and} \bibinfo{person}{Haifeng Wang}.} \bibinfo{year}{2022}\natexlab{}.
\newblock \showarticletitle{Geometry-enhanced molecular representation learning for property prediction}.
\newblock \bibinfo{journal}{\emph{Nature Machine Intelligence}} \bibinfo{volume}{4}, \bibinfo{number}{2} (\bibinfo{year}{2022}), \bibinfo{pages}{127--134}.
\newblock


\bibitem[Finn et~al\mbox{.}(2017)]%
        {finn2017model}
\bibfield{author}{\bibinfo{person}{Chelsea Finn}, \bibinfo{person}{Pieter Abbeel}, {and} \bibinfo{person}{Sergey Levine}.} \bibinfo{year}{2017}\natexlab{}.
\newblock \showarticletitle{Model-agnostic meta-learning for fast adaptation of deep networks}. In \bibinfo{booktitle}{\emph{International conference on machine learning}}. PMLR, \bibinfo{pages}{1126--1135}.
\newblock


\bibitem[Gaulton et~al\mbox{.}(2012)]%
        {gaulton2012chembl}
\bibfield{author}{\bibinfo{person}{Anna Gaulton}, \bibinfo{person}{Louisa~J Bellis}, \bibinfo{person}{A~Patricia Bento}, \bibinfo{person}{Jon Chambers}, \bibinfo{person}{Mark Davies}, \bibinfo{person}{Anne Hersey}, \bibinfo{person}{Yvonne Light}, \bibinfo{person}{Shaun McGlinchey}, \bibinfo{person}{David Michalovich}, \bibinfo{person}{Bissan Al-Lazikani}, {et~al\mbox{.}}} \bibinfo{year}{2012}\natexlab{}.
\newblock \showarticletitle{ChEMBL: a large-scale bioactivity database for drug discovery}.
\newblock \bibinfo{journal}{\emph{Nucleic acids research}} \bibinfo{volume}{40}, \bibinfo{number}{D1} (\bibinfo{year}{2012}), \bibinfo{pages}{D1100--D1107}.
\newblock


\bibitem[Gilmer et~al\mbox{.}(2017)]%
        {gilmer2017neural}
\bibfield{author}{\bibinfo{person}{Justin Gilmer}, \bibinfo{person}{Samuel~S Schoenholz}, \bibinfo{person}{Patrick~F Riley}, \bibinfo{person}{Oriol Vinyals}, {and} \bibinfo{person}{George~E Dahl}.} \bibinfo{year}{2017}\natexlab{}.
\newblock \showarticletitle{Neural message passing for quantum chemistry}. In \bibinfo{booktitle}{\emph{International conference on machine learning}}. PMLR, \bibinfo{pages}{1263--1272}.
\newblock


\bibitem[Glorot and Bengio(2010)]%
        {pmlr-v9-glorot10a}
\bibfield{author}{\bibinfo{person}{Xavier Glorot} {and} \bibinfo{person}{Yoshua Bengio}.} \bibinfo{year}{2010}\natexlab{}.
\newblock \showarticletitle{Understanding the difficulty of training deep feedforward neural networks}. In \bibinfo{booktitle}{\emph{Proceedings of the Thirteenth International Conference on Artificial Intelligence and Statistics}} \emph{(\bibinfo{series}{Proceedings of Machine Learning Research}, Vol.~\bibinfo{volume}{9})}, \bibfield{editor}{\bibinfo{person}{Yee~Whye Teh} {and} \bibinfo{person}{Mike Titterington}} (Eds.). \bibinfo{publisher}{PMLR}, \bibinfo{pages}{249--256}.
\newblock


\bibitem[Guo et~al\mbox{.}(2020)]%
        {guo2020graseq}
\bibfield{author}{\bibinfo{person}{Zhichun Guo}, \bibinfo{person}{Wenhao Yu}, \bibinfo{person}{Chuxu Zhang}, \bibinfo{person}{Meng Jiang}, {and} \bibinfo{person}{Nitesh~V Chawla}.} \bibinfo{year}{2020}\natexlab{}.
\newblock \showarticletitle{GraSeq: graph and sequence fusion learning for molecular property prediction}. In \bibinfo{booktitle}{\emph{Proceedings of the 29th ACM international conference on information \& knowledge management}}. \bibinfo{pages}{435--443}.
\newblock


\bibitem[Guo et~al\mbox{.}(2021)]%
        {guo2021few}
\bibfield{author}{\bibinfo{person}{Zhichun Guo}, \bibinfo{person}{Chuxu Zhang}, \bibinfo{person}{Wenhao Yu}, \bibinfo{person}{John Herr}, \bibinfo{person}{Olaf Wiest}, \bibinfo{person}{Meng Jiang}, {and} \bibinfo{person}{Nitesh~V Chawla}.} \bibinfo{year}{2021}\natexlab{}.
\newblock \showarticletitle{Few-shot graph learning for molecular property prediction}. In \bibinfo{booktitle}{\emph{Proceedings of the web conference 2021}}. \bibinfo{pages}{2559--2567}.
\newblock


\bibitem[Hansen et~al\mbox{.}(2015)]%
        {hansen2015machine}
\bibfield{author}{\bibinfo{person}{Katja Hansen}, \bibinfo{person}{Franziska Biegler}, \bibinfo{person}{Raghunathan Ramakrishnan}, \bibinfo{person}{Wiktor Pronobis}, \bibinfo{person}{O~Anatole Von~Lilienfeld}, \bibinfo{person}{Klaus-Robert Muller}, {and} \bibinfo{person}{Alexandre Tkatchenko}.} \bibinfo{year}{2015}\natexlab{}.
\newblock \showarticletitle{Machine learning predictions of molecular properties: Accurate many-body potentials and nonlocality in chemical space}.
\newblock \bibinfo{journal}{\emph{The journal of physical chemistry letters}} \bibinfo{volume}{6}, \bibinfo{number}{12} (\bibinfo{year}{2015}), \bibinfo{pages}{2326--2331}.
\newblock


\bibitem[Hu et~al\mbox{.}(2019)]%
        {hu2019strategies}
\bibfield{author}{\bibinfo{person}{Weihua Hu}, \bibinfo{person}{Bowen Liu}, \bibinfo{person}{Joseph Gomes}, \bibinfo{person}{Marinka Zitnik}, \bibinfo{person}{Percy Liang}, \bibinfo{person}{Vijay Pande}, {and} \bibinfo{person}{Jure Leskovec}.} \bibinfo{year}{2019}\natexlab{}.
\newblock \showarticletitle{Strategies for pre-training graph neural networks}.
\newblock \bibinfo{journal}{\emph{arXiv preprint arXiv:1905.12265}} (\bibinfo{year}{2019}).
\newblock


\bibitem[Jin et~al\mbox{.}(2020)]%
        {jin2020hierarchical}
\bibfield{author}{\bibinfo{person}{Wengong Jin}, \bibinfo{person}{Regina Barzilay}, {and} \bibinfo{person}{Tommi Jaakkola}.} \bibinfo{year}{2020}\natexlab{}.
\newblock \showarticletitle{Hierarchical generation of molecular graphs using structural motifs}. In \bibinfo{booktitle}{\emph{International conference on machine learning}}. PMLR, \bibinfo{pages}{4839--4848}.
\newblock


\bibitem[Kim et~al\mbox{.}(2019)]%
        {kim2019edge}
\bibfield{author}{\bibinfo{person}{Jongmin Kim}, \bibinfo{person}{Taesup Kim}, \bibinfo{person}{Sungwoong Kim}, {and} \bibinfo{person}{Chang~D Yoo}.} \bibinfo{year}{2019}\natexlab{}.
\newblock \showarticletitle{Edge-labeling graph neural network for few-shot learning}. In \bibinfo{booktitle}{\emph{Proceedings of the IEEE/CVF conference on computer vision and pattern recognition}}. \bibinfo{pages}{11--20}.
\newblock


\bibitem[Kim et~al\mbox{.}(2023)]%
        {dlmpp3}
\bibfield{author}{\bibinfo{person}{Suyeon Kim}, \bibinfo{person}{Dongha Lee}, \bibinfo{person}{SeongKu Kang}, \bibinfo{person}{Seonghyeon Lee}, {and} \bibinfo{person}{Hwanjo Yu}.} \bibinfo{year}{2023}\natexlab{}.
\newblock \showarticletitle{Learning topology-specific experts for molecular property prediction}. In \bibinfo{booktitle}{\emph{Proceedings of the AAAI Conference on Artificial Intelligence}}, Vol.~\bibinfo{volume}{37}. \bibinfo{pages}{8291--8299}.
\newblock


\bibitem[Kipf and Welling(2017)]%
        {kipf2017gcn}
\bibfield{author}{\bibinfo{person}{Thomas~N. Kipf} {and} \bibinfo{person}{Max Welling}.} \bibinfo{year}{2017}\natexlab{}.
\newblock \showarticletitle{Semi-Supervised Classification with Graph Convolutional Networks}. In \bibinfo{booktitle}{\emph{International Conference on Learning Representations (ICLR)}}.
\newblock


\bibitem[Koch et~al\mbox{.}(2015)]%
        {koch2015siamese}
\bibfield{author}{\bibinfo{person}{Gregory Koch}, \bibinfo{person}{Richard Zemel}, \bibinfo{person}{Ruslan Salakhutdinov}, {et~al\mbox{.}}} \bibinfo{year}{2015}\natexlab{}.
\newblock \showarticletitle{Siamese neural networks for one-shot image recognition}. In \bibinfo{booktitle}{\emph{ICML deep learning workshop}}, Vol.~\bibinfo{volume}{2}. Lille, \bibinfo{pages}{1--30}.
\newblock


\bibitem[Liu et~al\mbox{.}(2018)]%
        {liu2018learning}
\bibfield{author}{\bibinfo{person}{Yanbin Liu}, \bibinfo{person}{Juho Lee}, \bibinfo{person}{Minseop Park}, \bibinfo{person}{Saehoon Kim}, \bibinfo{person}{Eunho Yang}, \bibinfo{person}{Sung~Ju Hwang}, {and} \bibinfo{person}{Yi Yang}.} \bibinfo{year}{2018}\natexlab{}.
\newblock \showarticletitle{Learning to propagate labels: Transductive propagation network for few-shot learning}.
\newblock \bibinfo{journal}{\emph{arXiv preprint arXiv:1805.10002}} (\bibinfo{year}{2018}).
\newblock


\bibitem[Lu et~al\mbox{.}(2019)]%
        {lu2019molecular}
\bibfield{author}{\bibinfo{person}{Chengqiang Lu}, \bibinfo{person}{Qi Liu}, \bibinfo{person}{Chao Wang}, \bibinfo{person}{Zhenya Huang}, \bibinfo{person}{Peize Lin}, {and} \bibinfo{person}{Lixin He}.} \bibinfo{year}{2019}\natexlab{}.
\newblock \showarticletitle{Molecular property prediction: A multilevel quantum interactions modeling perspective}. In \bibinfo{booktitle}{\emph{Proceedings of the AAAI conference on artificial intelligence}}, Vol.~\bibinfo{volume}{33}. \bibinfo{pages}{1052--1060}.
\newblock


\bibitem[Meng et~al\mbox{.}(2023)]%
        {meng2023meta}
\bibfield{author}{\bibinfo{person}{Ziqiao Meng}, \bibinfo{person}{Yaoman Li}, \bibinfo{person}{Peilin Zhao}, \bibinfo{person}{Yang Yu}, {and} \bibinfo{person}{Irwin King}.} \bibinfo{year}{2023}\natexlab{}.
\newblock \showarticletitle{Meta-learning with motif-based task augmentation for few-shot molecular property prediction}. In \bibinfo{booktitle}{\emph{Proceedings of the 2023 SIAM International Conference on Data Mining (SDM)}}. SIAM, \bibinfo{pages}{811--819}.
\newblock


\bibitem[Nguyen et~al\mbox{.}(2020)]%
        {nguyen2020meta}
\bibfield{author}{\bibinfo{person}{Cuong~Q Nguyen}, \bibinfo{person}{Constantine Kreatsoulas}, {and} \bibinfo{person}{Kim~M Branson}.} \bibinfo{year}{2020}\natexlab{}.
\newblock \showarticletitle{Meta-learning GNN initializations for low-resource molecular property prediction}.
\newblock \bibinfo{journal}{\emph{arXiv preprint arXiv:2003.05996}} (\bibinfo{year}{2020}).
\newblock


\bibitem[Riniker and Landrum(2013)]%
        {riniker2013similarity}
\bibfield{author}{\bibinfo{person}{Sereina Riniker} {and} \bibinfo{person}{Gregory~A Landrum}.} \bibinfo{year}{2013}\natexlab{}.
\newblock \showarticletitle{Similarity maps-a visualization strategy for molecular fingerprints and machine-learning methods}.
\newblock \bibinfo{journal}{\emph{Journal of cheminformatics}}  \bibinfo{volume}{5} (\bibinfo{year}{2013}), \bibinfo{pages}{1--7}.
\newblock


\bibitem[Schimunek et~al\mbox{.}(2023)]%
        {schimunek2023context}
\bibfield{author}{\bibinfo{person}{Johannes Schimunek}, \bibinfo{person}{Philipp Seidl}, \bibinfo{person}{Lukas Friedrich}, \bibinfo{person}{Daniel Kuhn}, \bibinfo{person}{Friedrich Rippmann}, \bibinfo{person}{Sepp Hochreiter}, {and} \bibinfo{person}{G{\""u}nter Klambauer}.} \bibinfo{year}{2023}\natexlab{}.
\newblock \showarticletitle{Context-enriched molecule representations improve few-shot drug discovery}.
\newblock \bibinfo{journal}{\emph{arXiv preprint arXiv:2305.09481}} (\bibinfo{year}{2023}).
\newblock


\bibitem[Sliwoski et~al\mbox{.}(2014)]%
        {sliwoski2014computational}
\bibfield{author}{\bibinfo{person}{Gregory Sliwoski}, \bibinfo{person}{Sandeepkumar Kothiwale}, \bibinfo{person}{Jens Meiler}, {and} \bibinfo{person}{Edward~W Lowe~Jr}.} \bibinfo{year}{2014}\natexlab{}.
\newblock \showarticletitle{Computational methods in drug discovery}.
\newblock \bibinfo{journal}{\emph{Pharmacological reviews}} \bibinfo{volume}{66}, \bibinfo{number}{1} (\bibinfo{year}{2014}), \bibinfo{pages}{334--395}.
\newblock


\bibitem[Snell et~al\mbox{.}(2017)]%
        {snell2017prototypical}
\bibfield{author}{\bibinfo{person}{Jake Snell}, \bibinfo{person}{Kevin Swersky}, {and} \bibinfo{person}{Richard Zemel}.} \bibinfo{year}{2017}\natexlab{}.
\newblock \showarticletitle{Prototypical networks for few-shot learning}.
\newblock \bibinfo{journal}{\emph{Advances in neural information processing systems}}  \bibinfo{volume}{30} (\bibinfo{year}{2017}).
\newblock


\bibitem[Song et~al\mbox{.}(2020)]%
        {dlmpp6}
\bibfield{author}{\bibinfo{person}{Ying Song}, \bibinfo{person}{Shuangjia Zheng}, \bibinfo{person}{Zhangming Niu}, \bibinfo{person}{Zhang-Hua Fu}, \bibinfo{person}{Yutong Lu}, {and} \bibinfo{person}{Yuedong Yang}.} \bibinfo{year}{2020}\natexlab{}.
\newblock \showarticletitle{Communicative representation learning on attributed molecular graphs}. In \bibinfo{booktitle}{\emph{29th International Joint Conference on Artificial Intelligence and the 17th Pacific Rim International Conference on Artificial Intelligence (IJCAI-PRICAI2020)}}. International Joint Conferences on Artificial Intelligence Organization.
\newblock


\bibitem[Sun et~al\mbox{.}(2021)]%
        {sun2021sugar}
\bibfield{author}{\bibinfo{person}{Qingyun Sun}, \bibinfo{person}{Jianxin Li}, \bibinfo{person}{Hao Peng}, \bibinfo{person}{Jia Wu}, \bibinfo{person}{Yuanxing Ning}, \bibinfo{person}{Philip~S Yu}, {and} \bibinfo{person}{Lifang He}.} \bibinfo{year}{2021}\natexlab{}.
\newblock \showarticletitle{Sugar: Subgraph neural network with reinforcement pooling and self-supervised mutual information mechanism}. In \bibinfo{booktitle}{\emph{Proceedings of the web conference 2021}}. \bibinfo{pages}{2081--2091}.
\newblock


\bibitem[Tang et~al\mbox{.}(2020)]%
        {tang2020investigating}
\bibfield{author}{\bibinfo{person}{Xianfeng Tang}, \bibinfo{person}{Huaxiu Yao}, \bibinfo{person}{Yiwei Sun}, \bibinfo{person}{Yiqi Wang}, \bibinfo{person}{Jiliang Tang}, \bibinfo{person}{Charu Aggarwal}, \bibinfo{person}{Prasenjit Mitra}, {and} \bibinfo{person}{Suhang Wang}.} \bibinfo{year}{2020}\natexlab{}.
\newblock \showarticletitle{Investigating and mitigating degree-related biases in graph convoltuional networks}. In \bibinfo{booktitle}{\emph{Proceedings of the 29th ACM International Conference on Information \& Knowledge Management}}. \bibinfo{pages}{1435--1444}.
\newblock


\bibitem[Ullmann(1976)]%
        {ullmann1976algorithm}
\bibfield{author}{\bibinfo{person}{Julian~R Ullmann}.} \bibinfo{year}{1976}\natexlab{}.
\newblock \showarticletitle{An algorithm for subgraph isomorphism}.
\newblock \bibinfo{journal}{\emph{Journal of the ACM (JACM)}} \bibinfo{volume}{23}, \bibinfo{number}{1} (\bibinfo{year}{1976}), \bibinfo{pages}{31--42}.
\newblock


\bibitem[Wang et~al\mbox{.}(2024)]%
        {wang2024pin}
\bibfield{author}{\bibinfo{person}{Liang Wang}, \bibinfo{person}{Qiang Liu}, \bibinfo{person}{Shaozhen Liu}, \bibinfo{person}{Xin Sun}, {and} \bibinfo{person}{Shu Wu}.} \bibinfo{year}{2024}\natexlab{}.
\newblock \showarticletitle{Pin-tuning: Parameter-efficient in-context tuning for few-shot molecular property prediction}.
\newblock \bibinfo{journal}{\emph{Advances in Neural Information Processing Systems}}  \bibinfo{volume}{37} (\bibinfo{year}{2024}), \bibinfo{pages}{70765--70792}.
\newblock


\bibitem[Wang et~al\mbox{.}(2023)]%
        {dlmpp5}
\bibfield{author}{\bibinfo{person}{Xu Wang}, \bibinfo{person}{Huan Zhao}, \bibinfo{person}{Wei-wei Tu}, {and} \bibinfo{person}{Quanming Yao}.} \bibinfo{year}{2023}\natexlab{}.
\newblock \showarticletitle{Automated 3d pre-training for molecular property prediction}. In \bibinfo{booktitle}{\emph{Proceedings of the 29th ACM SIGKDD Conference on Knowledge Discovery and Data Mining}}. \bibinfo{pages}{2419--2430}.
\newblock


\bibitem[Wang et~al\mbox{.}(2021)]%
        {wang2021property}
\bibfield{author}{\bibinfo{person}{Yaqing Wang}, \bibinfo{person}{Abulikemu Abuduweili}, \bibinfo{person}{Quanming Yao}, {and} \bibinfo{person}{Dejing Dou}.} \bibinfo{year}{2021}\natexlab{}.
\newblock \showarticletitle{Property-aware relation networks for few-shot molecular property prediction}.
\newblock \bibinfo{journal}{\emph{Advances in Neural Information Processing Systems}}  \bibinfo{volume}{34} (\bibinfo{year}{2021}), \bibinfo{pages}{17441--17454}.
\newblock


\bibitem[Wu et~al\mbox{.}(2023)]%
        {wu2023molformer}
\bibfield{author}{\bibinfo{person}{Fang Wu}, \bibinfo{person}{Dragomir Radev}, {and} \bibinfo{person}{Stan~Z Li}.} \bibinfo{year}{2023}\natexlab{}.
\newblock \showarticletitle{Molformer: Motif-based transformer on 3d heterogeneous molecular graphs}. In \bibinfo{booktitle}{\emph{Proceedings of the AAAI Conference on Artificial Intelligence}}, Vol.~\bibinfo{volume}{37}. \bibinfo{pages}{5312--5320}.
\newblock


\bibitem[Wu et~al\mbox{.}(2024)]%
        {frag1-wu2024t}
\bibfield{author}{\bibinfo{person}{Juan-Ni Wu}, \bibinfo{person}{Tong Wang}, \bibinfo{person}{Yue Chen}, \bibinfo{person}{Li-Juan Tang}, \bibinfo{person}{Hai-Long Wu}, {and} \bibinfo{person}{Ru-Qin Yu}.} \bibinfo{year}{2024}\natexlab{}.
\newblock \showarticletitle{t-SMILES: a fragment-based molecular representation framework for de novo ligand design}.
\newblock \bibinfo{journal}{\emph{Nature Communications}} \bibinfo{volume}{15}, \bibinfo{number}{1} (\bibinfo{year}{2024}), \bibinfo{pages}{4993}.
\newblock


\bibitem[Wu et~al\mbox{.}(2018)]%
        {wu2018moleculenet}
\bibfield{author}{\bibinfo{person}{Zhenqin Wu}, \bibinfo{person}{Bharath Ramsundar}, \bibinfo{person}{Evan~N Feinberg}, \bibinfo{person}{Joseph Gomes}, \bibinfo{person}{Caleb Geniesse}, \bibinfo{person}{Aneesh~S Pappu}, \bibinfo{person}{Karl Leswing}, {and} \bibinfo{person}{Vijay Pande}.} \bibinfo{year}{2018}\natexlab{}.
\newblock \showarticletitle{MoleculeNet: a benchmark for molecular machine learning}.
\newblock \bibinfo{journal}{\emph{Chemical science}} \bibinfo{volume}{9}, \bibinfo{number}{2} (\bibinfo{year}{2018}), \bibinfo{pages}{513--530}.
\newblock


\bibitem[Yang et~al\mbox{.}(2021)]%
        {yang2021graphformers}
\bibfield{author}{\bibinfo{person}{Junhan Yang}, \bibinfo{person}{Zheng Liu}, \bibinfo{person}{Shitao Xiao}, \bibinfo{person}{Chaozhuo Li}, \bibinfo{person}{Defu Lian}, \bibinfo{person}{Sanjay Agrawal}, \bibinfo{person}{Amit Singh}, \bibinfo{person}{Guangzhong Sun}, {and} \bibinfo{person}{Xing Xie}.} \bibinfo{year}{2021}\natexlab{}.
\newblock \showarticletitle{Graphformers: Gnn-nested transformers for representation learning on textual graph}.
\newblock \bibinfo{journal}{\emph{Advances in Neural Information Processing Systems}}  \bibinfo{volume}{34} (\bibinfo{year}{2021}), \bibinfo{pages}{28798--28810}.
\newblock


\bibitem[Yu and Gao(2022)]%
        {yu2022molecular}
\bibfield{author}{\bibinfo{person}{Zhaoning Yu} {and} \bibinfo{person}{Hongyang Gao}.} \bibinfo{year}{2022}\natexlab{}.
\newblock \showarticletitle{Molecular representation learning via heterogeneous motif graph neural networks}. In \bibinfo{booktitle}{\emph{International Conference on Machine Learning}}. PMLR, \bibinfo{pages}{25581--25594}.
\newblock


\bibitem[Zang et~al\mbox{.}(2023)]%
        {frag2-zang2023hierarchical}
\bibfield{author}{\bibinfo{person}{Xuan Zang}, \bibinfo{person}{Xianbing Zhao}, {and} \bibinfo{person}{Buzhou Tang}.} \bibinfo{year}{2023}\natexlab{}.
\newblock \showarticletitle{Hierarchical molecular graph self-supervised learning for property prediction}.
\newblock \bibinfo{journal}{\emph{Communications Chemistry}} \bibinfo{volume}{6}, \bibinfo{number}{1} (\bibinfo{year}{2023}), \bibinfo{pages}{34}.
\newblock


\bibitem[Zhang et~al\mbox{.}(2021a)]%
        {frag1-zhang2021fragat}
\bibfield{author}{\bibinfo{person}{Ziqiao Zhang}, \bibinfo{person}{Jihong Guan}, {and} \bibinfo{person}{Shuigeng Zhou}.} \bibinfo{year}{2021}\natexlab{a}.
\newblock \showarticletitle{FraGAT: a fragment-oriented multi-scale graph attention model for molecular property prediction}.
\newblock \bibinfo{journal}{\emph{Bioinformatics}} \bibinfo{volume}{37}, \bibinfo{number}{18} (\bibinfo{year}{2021}), \bibinfo{pages}{2981--2987}.
\newblock


\bibitem[Zhang et~al\mbox{.}(2021b)]%
        {zhang2021motif}
\bibfield{author}{\bibinfo{person}{Zaixi Zhang}, \bibinfo{person}{Qi Liu}, \bibinfo{person}{Hao Wang}, \bibinfo{person}{Chengqiang Lu}, {and} \bibinfo{person}{Chee-Kong Lee}.} \bibinfo{year}{2021}\natexlab{b}.
\newblock \showarticletitle{Motif-based graph self-supervised learning for molecular property prediction}.
\newblock \bibinfo{journal}{\emph{Advances in Neural Information Processing Systems}}  \bibinfo{volume}{34} (\bibinfo{year}{2021}), \bibinfo{pages}{15870--15882}.
\newblock


\bibitem[Zhou et~al\mbox{.}(2024)]%
        {zhou2024slotgat}
\bibfield{author}{\bibinfo{person}{Ziang Zhou}, \bibinfo{person}{Jieming Shi}, \bibinfo{person}{Renchi Yang}, \bibinfo{person}{Yuanhang Zou}, {and} \bibinfo{person}{Qing Li}.} \bibinfo{year}{2024}\natexlab{}.
\newblock \showarticletitle{SlotGAT: Slot-based Message Passing for Heterogeneous Graph Neural Network}.
\newblock \bibinfo{journal}{\emph{arXiv preprint arXiv:2405.01927}} (\bibinfo{year}{2024}).
\newblock


\bibitem[Zhu et~al\mbox{.}(2022)]%
        {frag2-zhu2022hignn}
\bibfield{author}{\bibinfo{person}{Weimin Zhu}, \bibinfo{person}{Yi Zhang}, \bibinfo{person}{Duancheng Zhao}, \bibinfo{person}{Jianrong Xu}, {and} \bibinfo{person}{Ling Wang}.} \bibinfo{year}{2022}\natexlab{}.
\newblock \showarticletitle{HiGNN: A hierarchical informative graph neural network for molecular property prediction equipped with feature-wise attention}.
\newblock \bibinfo{journal}{\emph{Journal of Chemical Information and Modeling}} \bibinfo{volume}{63}, \bibinfo{number}{1} (\bibinfo{year}{2022}), \bibinfo{pages}{43--55}.
\newblock


\bibitem[Zhuang et~al\mbox{.}(2023)]%
        {zhuang2023graph}
\bibfield{author}{\bibinfo{person}{Xiang Zhuang}, \bibinfo{person}{Qiang Zhang}, \bibinfo{person}{Bin Wu}, \bibinfo{person}{Keyan Ding}, \bibinfo{person}{Yin Fang}, {and} \bibinfo{person}{Huajun Chen}.} \bibinfo{year}{2023}\natexlab{}.
\newblock \showarticletitle{Graph sampling-based meta-learning for molecular property prediction}.
\newblock \bibinfo{journal}{\emph{arXiv preprint arXiv:2306.16780}} (\bibinfo{year}{2023}).
\newblock


\end{thebibliography}
\newpage

\begin{table*}[t!]
\centering
\caption{5-shot performance on ToxCast benchmarks}
\label{tab:toxcast-5shot}
{\footnotesize
\centering
\begin{tabularx}{\linewidth}{l|Y|Y|Y|Y|Y|Y|Y|Y|Y}
\toprule
\textbf{Method} & \textbf{APR} & \textbf{ATG} & \textbf{BSK} & \textbf{CEETOX} & \textbf{CLD} & \textbf{NVS} & \textbf{OT} & \textbf{TOX21} & \textbf{Tanguay} \\
\midrule
ProtoNet & 70.38 & 58.11 & 63.96 & 63.41 & 76.70 & 62.27 & 64.52 & 65.99 & 70.98 \\
MAML & 68.88 & 60.01 & 67.05 & 62.42 & 73.32 & 69.18 & 64.56 & 66.73 & 75.88 \\
TPN & 70.76 & 57.92 & 63.41 & 64.73 & 70.44 & 61.36 & 61.99 & 66.49 & 77.27 \\
EGNN & 74.06 & 60.56 & 64.60 & 63.20 & 71.44 & 62.62 & 66.70 & 65.33 & 75.69 \\
\midrule
Pre-GNN & 80.38 & 66.96 & 75.64 & 64.88 & 78.03 & 74.08 & 70.42 & 75.74 & 82.73 \\
Meta-MGNN & 81.22 & 69.90 & 79.67 & 65.78 & 77.53 & 73.99 & 69.20 & 76.25 & 83.76 \\
PAR & 83.76 & 70.24 & 80.82 & 69.51 & 81.32 & 70.60 & 71.31 & 79.71 & 84.71 \\
GS-Meta & 89.36 & 81.92 & 86.12 & 74.48 & 83.10 & 74.72 & 73.26 & 89.71 & 91.15 \\
Pin-Tuning & 89.94 & 82.37 & 87.61 & 75.20 & 85.07 & 75.49 & 74.70 & 90.89 & 92.14  \\
M-GLC (Ours) & \textbf{96.46} &
\textbf{90.59} & \textbf{92.58} & 
\textbf{82.56} & \textbf{90.25} & 
\textbf{77.60} & \textbf{88.47} & 
\textbf{95.33} & \textbf{99.36} \\
\bottomrule
\end{tabularx}
}
\end{table*}

\begin{table*}[htbp]
\centering
\caption{10-shot performance on ToxCast benchmarks}
\label{tab:toxcast-10shot}
{\footnotesize
\begin{tabularx}{\linewidth}{l|Y|Y|Y|Y|Y|Y|Y|Y|Y}
\toprule
\textbf{Method} & \textbf{APR} & \textbf{ATG} & \textbf{BSK} & \textbf{CEETOX} & \textbf{CLD} & \textbf{NVS} & \textbf{OT} & \textbf{TOX21} & \textbf{Tanguay} \\
\midrule
ProtoNet & 73.58 & 59.26 & 70.15 & 66.12 & 78.12 & 65.85 & 64.90 & 68.26 & 73.61\\
MAML & 72.66 & 62.09 & 66.42 & 64.08 & 74.57 & 66.56 & 64.07 & 68.04 & 77.12\\
TPN & 74.53 & 60.74 & 65.19 & 66.63 & 75.22 & 63.20 & 64.63 & 73.30 & 81.75\\
EGNN & 80.33 & 66.17 & 73.43 & 66.51 & 78.85 & 71.05 & 68.21 & 76.40 & 85.23\\
\midrule
Pre-GNN & 80.61 & 67.59 & 76.65 & 66.52 & 78.88 & 75.09 & 70.52 & 77.92 & 83.05  \\
Meta-MGNN & 81.47 & 69.20 & 78.97 & 66.57 & 78.30 & \textbf{79.60} & 69.55 & 78.77 & 83.98  \\
PAR & 86.09 & 72.72 & 82.45 & 72.12 & 83.43 & 74.94 & 71.96 & 82.81 & 88.20  \\
GS-Meta & 90.15 & 82.54 & 88.21 & 74.19 & 86.34 & 76.29 & 74.47 & 90.63 & 91.47  \\
Pin-Tuning &  92.78 &  83.58 &  89.49 &  75.96 &  87.70 &  76.33 &  75.56 &  90.80 &  92.25  \\
M-GLC (Ours) & \textbf{96.22} &
\textbf{92.24} & \textbf{92.57} & 
\textbf{81.58} & \textbf{89.33} & 
77.86 & \textbf{87.45} & 
\textbf{95.77} & \textbf{99.15} \\
\bottomrule
\end{tabularx}
}
\end{table*}

\appendix
\begin{figure}[t]
  \centering
  \includegraphics[width=1.0\linewidth]{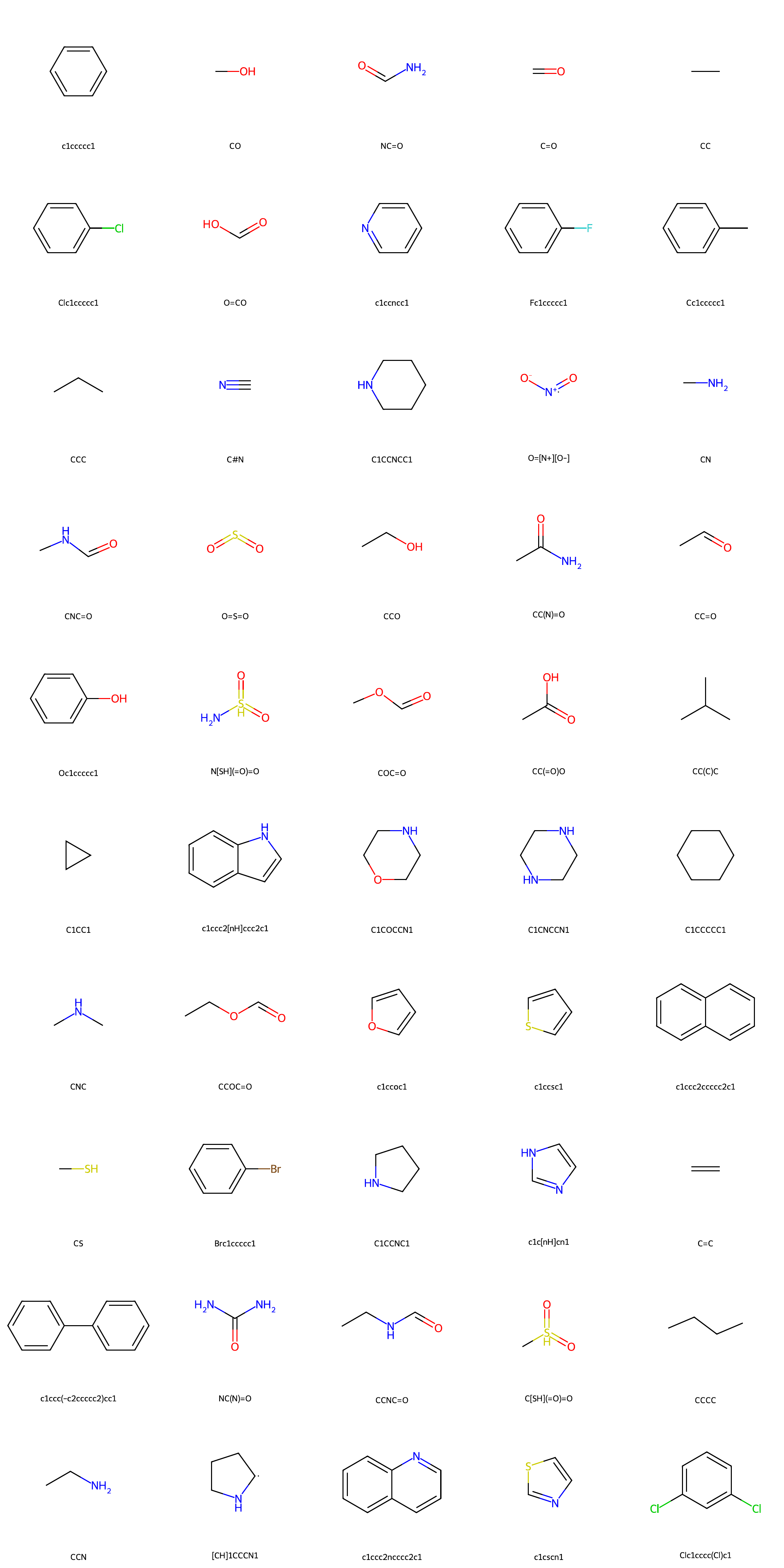} 
  \Description{}
    \caption{
    Visualization of the 50 most frequent motifs in the ChEMBL dataset \citep{gaulton2012chembl}, extracted using the bridge method proposed in \citet{jin2020hierarchical}.
    }
  \label{motif-50}
\end{figure}

\section{Motif Visualization}\label{visualization}
To better understand the structural distribution of chemical motifs, we visualize the 50 most frequent substructures in the ChEMBL dataset \citep{gaulton2012chembl}. These motifs are extracted using the bridge method introduced in \citet{jin2020hierarchical}, which identifies common and potentially meaningful substructures that may contribute to molecular property prediction.

\section{ToxCast Experimental Result}\label{app:toxcast}
We report per‑subset AUC‑ROC on ToxCast in Table~\ref{tab:toxcast-5shot} and~\ref{tab:toxcast-10shot}. In the 5‑shot setting, our method achieves the best score on every subset. In the 10‑shot setting, it is best on all but one subset, where it is comparable (second best). All baseline results are taken from \citet{wang2024pin} and are means over 10 random seeds (0–9).

\end{document}